\documentclass[final,5p,times]{elsarticle}
\usepackage{amssymb}
\usepackage{amsmath}
\usepackage[utf8x]{inputenc} 
\usepackage{algorithmic}
\usepackage[table,xcdraw]{xcolor}
\usepackage{float}
\usepackage[shortlabels,inline]{enumitem}
\usepackage{makecell}

\usepackage{subcaption}
\usepackage{longtable}
\usepackage{multirow}
\usepackage{url}
\usepackage{pifont}
\usepackage{textcomp}
\usepackage{ragged2e}
\usepackage[linesnumbered,ruled,vlined]{algorithm2e}
\usepackage{bm}
\usepackage{arydshln}
\usepackage{multirow}
\usepackage{tabularray}
\usepackage{float}
\usepackage{colortbl}
\usepackage{booktabs}
\usepackage{siunitx}
\usepackage{hhline}
\usepackage{orcidlink}
\usepackage[font=small,justification=justified,format=plain,skip=0pt,belowskip=0pt,aboveskip=0pt]{caption}
\usepackage{tcolorbox}
\usepackage[normalem]{ulem}
\usepackage{tikz}
\usepackage[capitalise,nameinlink]{cleveref}
\usepackage{amsmath}
\usepackage{etoolbox}
\usepackage{marginnote}
\usepackage{mdframed}

\newfloat{infobox}{tbp}{lob}
\floatname{infobox}{Box}

\newmdenv[
  linewidth=0.8pt,
  roundcorner=3mm,
  shadow=true,
  shadowsize=1pt,
  backgroundcolor=blue!3!white,
  linecolor=blue!70!black,
  innertopmargin=8pt,
  innerbottommargin=8pt,
  innerleftmargin=5pt,
  innerrightmargin=5pt,
  frametitlefont=\sffamily\color{white},
  frametitlebackgroundcolor=blue!75!black,
  frametitleaboveskip=1pt,
  frametitlebelowskip=1pt,
  frametitlealignment=\centering,
  nobreak=true
]{mybox}

\newmdenv[
  linewidth=0.8pt,
  roundcorner=3mm,
  shadow=true,
  shadowsize=5pt,
  backgroundcolor=red!3!white,
  linecolor=red!70!black,
  innertopmargin=8pt,
  innerbottommargin=8pt,
  innerleftmargin=5pt,
  innerrightmargin=5pt,
  frametitlefont=\sffamily\color{white},
  frametitlebackgroundcolor=red!75!black,
  frametitleaboveskip=1pt,
  frametitlebelowskip=1pt,
  frametitlealignment=\centering,
  nobreak=true
]{redbox}

\reversemarginpar 

\makeatletter
\patchcmd{\@floatboxreset}{\@fs@pre}{\@fs@pre\setlength{\textfloatsep}{0pt}}{}{}
\patchcmd{\@floatboxreset}{\@fs@post}{\@fs@post\setlength{\textfloatsep}{0pt}}{}{}
\makeatother

\newcommand*\circled[1]{\tikz[baseline=(char.base)]{
            \node[shape=circle,draw,inner sep=2pt] (char) {#1};}}

\journal{Knowledge-Based Systems}
\begin{document}

\begin{frontmatter}

%% Title, authors and addresses

%% use the tnoteref command within \title for footnotes;
%% use the tnotetext command for theassociated footnote;
%% use the fnref command within \author or \affiliation for footnotes;
%% use the fntext command for theassociated footnote;
%% use the corref command within \author for corresponding author footnotes;
%% use the cortext command for theassociated footnote;
%% use the ead command for the email address,
%% and the form \ead[url] for the home page:
%% \title{Title\tnoteref{label1}}
%% \tnotetext[label1]{}
%% \author{Name\corref{cor1}\fnref{label2}}
%% \ead{email address}
%% \ead[url]{home page}
%% \fntext[label2]{}
%% \cortext[cor1]{}
%% \affiliation{organization={},
%%             addressline={},
%%             city={},
%%             postcode={},
%%             state={},
%%             country={}}
%% \fntext[label3]{}

\title{AutoStreamPipe: LLM Assisted Automatic Generation of Data Stream Processing Pipelines}

% use optional labels to link authors explicitly to addresses:

\author{Abolfazl Younesi\orcidlink{0009-0003-0052-6475}\corref{cor1}\fnref{label1}}
\ead{abolfazl.younesi@uibk.ac.at }

\author{Zahra Najafabadi Samani\orcidlink{0000-0001-5182-9087}\corref{cor1}\fnref{label1}}
\ead{zahra.najafabadisamani@uibk.ac.at }

\author{Thomas Fahringer\orcidlink{0000-0003-4293-1228}\corref{cor1}\fnref{label1}}
\ead{thomas.fahringer@uibk.ac.at }

\cortext[cor1]{Corresponding authors}

\affiliation{organization={Departement of Computer Science, University of Innsbruck},
addressline={Technikerstra\ss e},
city={Innsbruck},
postcode={6020},
state={Tirol},
country={Austria}}

% \author[label1]{Abolfazl Younesi\orcidlink{0009-0003-0052-6475}}
% \author[label1]{Zahra Najafabadi Samani\orcidlink{0000-0001-5182-9087}}
% \author[label1]{Thomas Fahringer\orcidlink{0000-0003-4293-1228}}

% \affiliation[label1]{organization={Department of Computer Science, University of Innsbruck},
% addressline={Technikerstra\ss e},
% city={Innsbruck},
% postcode={6020},
% state={Tirol},
% country={Austria}}
% \author{} %% Author name

% %% Author affiliation
% \affiliation{organization={},%Department and Organization
%             addressline={}, 
%             city={},
%             postcode={}, 
%             state={},
%             country={}}

%% Abstract
\begin{abstract}
Data pipelines are essential in stream processing as they enable the efficient collection, processing, and delivery of real-time data, supporting rapid data analysis. In this paper, we present AutoStreamPipe, a novel framework that employs Large Language Models (LLMs) to automate the design, generation, and deployment of stream processing pipelines. AutoStreamPipe bridges the semantic gap between high-level user intent and platform-specific implementations across distributed stream processing systems for structured multi-agent reasoning by integrating a Hypergraph of Thoughts (HGoT) as an extended version of GoT. AutoStreamPipe combines resilient execution strategies, advanced query analysis, and HGoT to deliver pipelines with good accuracy. Experimental evaluations on diverse pipelines demonstrate that AutoStreamPipe significantly reduces development time (\texttimes 6.3) and error rates (\texttimes 5.19), as measured by a novel Error-Free Score (EFS), compared to LLM code-generation methods.
\end{abstract}

% %%Graphical abstract
% \begin{graphicalabstract}
% \includegraphics[width=2\columnwidth]{Figures/AutoPipe-details_new.drawio.pdf}

% \end{graphicalabstract}

% %%Research highlights
% \begin{highlights}
% \item AutoStreamPipe automates the full pipeline lifecycle from intent to deployment
% \item Hypergraph of Thoughts enables consistent multi-step pipeline reasoning
% \item Query analyzer extracts user intent and formalizes pipeline specifications
% \item Resilient multi-agent execution ensures fault-tolerant pipeline generation
% \item Achieves 6.3$\times$ faster development and 5.2$\times$ fewer errors vs. LLM baselines
% \end{highlights}

%% Keywords
\begin{keyword}
Large Language Models\sep Stream Processing\sep Data Pipelines\sep 
Workflow Automation\sep Hypergraph of Thoughts\sep Multi-Agent Systems\sep 
Pipeline Generation
\end{keyword}

\end{frontmatter}

\section{Introduction} \label{sec:introduction}
The rapid evolution of technology has made stream data processing essential rather than optional \cite{noghabi2017samza,cardellini2022runtime}. Stream processing (SP) pipelines form the backbone of systems that require low latency and high throughput for data. These systems range from monitoring IoT devices and sensor networks in cyber-physical systems to detecting fraudulent financial transactions \cite{9910406,10820033,9356167}. To meet these demands, SP pipelines must transform raw data streams into actionable insights in real time. However, their development remains complex and time-consuming due to the intricacies of distributed, stateful computation.

\textbf{Challenges in Pipeline Development.} Designing and deploying SP pipelines presents significant challenges. Traditional approaches typically involve manual coding, iterative debugging, and labor-intensive optimization \cite{akidau2015dataflow}. 

Developing data pipelines demands expertise in two key areas: domain-specific logic (e.g., rules for anomaly detection) and framework-specific APIs (e.g., Flink’s DataStream API), which poses a significant challenge for domain experts, such as data analysts, who may lack advanced programming skills. As a result, manual coding by engineers often involves considerable effort refining logic, debugging edge cases, and optimizing resource usage. 
Even automated code generation tools fall short, as they fail to bridge the semantic gap between high-level intent (e.g., "detect three consecutive failed logins within five minutes") and low-level implementation details (e.g., configuring Flink operators with custom triggers). This gap highlights the need for a paradigm that combines the accessibility of low-code interfaces with the expressivity of handcrafted code.

Although automated code generation and template-driven tools exist \cite{node-red,apache_nifi}, they often fail to address dynamic operations, particularly stateful operations \cite{akidau2015dataflow}. While effective for building pipelines with simple, common operators, they typically require users to write custom code for complex or domain-specific logic, reintroducing the manual effort they were designed to avoid.

To address these challenges, several stream processing application benchmarks have been proposed. However, existing benchmarks still exhibit significant limitations\cite{tucker2008nexmark, garcia2023spbench, shukla2017riotbench, bordin2020dspbench, hesse2021espbench, huang2010hibench, arasu2004linear, chintapalli2016benchmarking, li2015sparkbench, lu2014stream, wang2014bigdatabench, 10.1145/2463676.2463712, 10.1145/2463676.2465296, agnihotri2025pdspbenchbenchmarkingparalleldistributed}. Benchmarks such as RIoTBench \cite{shukla2017riotbench}, DSPBench \cite{bordin2020dspbench}, HiBench \cite{huang2010hibench}, BigDataBench \cite{wang2014bigdatabench}, the Linear Road Benchmark \cite{arasu2004linear}, NexMark \cite{tucker2008nexmark}, BigBench \cite{10.1145/2463676.2463712}, and LinkBench \cite{10.1145/2463676.2465296} often rely on outdated DSPS versions, support only a limited range of platforms, and fail to incorporate recent advancements in state management and event-time processing. As a result, their relevance to modern deployments is significantly constrained.
As shown in Table \ref{tab:comparison}, existing benchmarks rarely provide comprehensive cross-platform evaluations. Additionally, they typically offer a narrow set of predefined pipelines, many of which are synthetic and fail to reflect real-world scenarios.
These gaps in evaluation capabilities motivate our work. Rather than proposing another static benchmark suite, we focus on automating the generation of diverse SP pipelines, providing a flexible source of workloads that complements existing benchmarking efforts.

\textbf{Introducing LLMs for Pipeline Generation}. Recent advances in automation, driven by the rise of LLMs such as GPT-4 \cite{achiam2023gpt}, LLaMA \cite{touvron2023llama}, and Code Llama \cite{roziere2023code}, offer promising avenues to address pipeline development challenges. LLMs can understand natural language descriptions, generate code, and reason through complex logic. This could make advanced technologies, such as distributed stream processing systems like Apache Flink \cite{apache_flink, carbone2015apache}, Apache Storm \cite{apache_storm, toshniwal2014storm}, Kafka Streams \cite{apache_kafka}, and Apache Spark Streaming \cite{apache_spark}, more accessible.

\begin{figure}[b]
    \centering
    \includegraphics[width=1\linewidth]{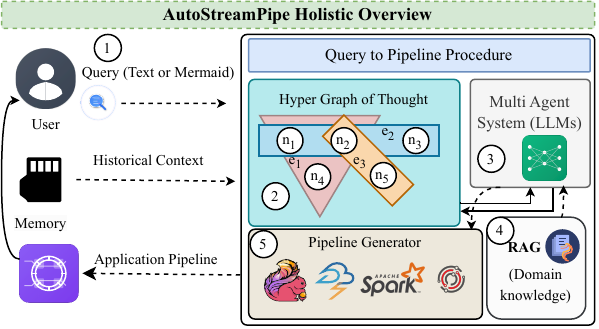}
    \caption{A holistic overview of the \textsc{AutoStreamPipe} framework}
    \label{fig:AutoStreamPipe_overview}
\end{figure}

Despite their potential, LLMs face significant limitations when applied to pipeline generation, including challenges with semantic accuracy, inconsistency in multi-step reasoning, and difficulty adapting to evolving requirements, such as changes in data schemas or real-time processing constraints. 
\textbf{Proposed Approach.} To address the challenge of rapid pipeline prototyping, we propose \textsc{AutoStreamPipe}, a framework that employs LLMs \cite{qu2025tool} to automate SP pipeline generation (cf. Figure~\ref{fig:AutoStreamPipe_overview} \circled{1}-\circled{5}). \textsc{AutoStreamPipe} includes a query analyzer integrated with the plan designer and short-term and long-term memory for future references. We embed a Hypergraph of Thought (HGoT) and integrate it with multiple LLM agents in a RAG setup, based on standard SP pipeline architectures, to produce production-ready pipelines. In contrast to general-purpose code generators that often produce unclear or non-optimizable execution plans using outdated or deprecated functions, we automate the creation of clean, high-level code. This approach ensures the framework serves as a solid starting point for quick validation and iteration with SP technologies. Our framework is capable of generating a virtually unlimited number of SP pipelines, supporting both synthetic and real-world use cases across all major DSPSs. This paradigm accelerates development cycles and lowers the technical barrier to using SP technologies, enabling domain experts without programming expertise to translate their specialized knowledge into operational systems.

\textbf{Contribution.} Our main contributions are as follows:
\begin{itemize}[wide]
    \item \textbf{End-to-End Automation:} Our system automates the entire pipeline lifecycle, from interpreting user inputs to generating, optimizing, and validating the final pipeline for DSPS based on high-level semantic understanding (cf.~\cref{sec:architecture}).
    \item \textbf{Query Analyzer}: We introduce a dedicated module for intent detection and parameter extraction that deconstructs natural language queries into formal pipeline specifications. It performs deep semantic analysis to infer implicit constraints, validate explicit user constraints, and generate execution plans tailored to varying pipeline complexities (cf.~\cref{sec:Query_analysis}).
    \item \textbf{Hypergraph of Thoughts (HGoT) Reasoning:} We propose HGoT, a novel structured reasoning framework that extends the Graph of Thoughts (GoT) paradigm~\cite{besta2024graph} by introducing hyperedges to model multi-way dependencies among partial solutions. HGoT enables coordinated reasoning across interdependent steps, improving consistency and efficiency in complex pipeline synthesis tasks (cf.~\cref{subsec: Graph Building}).
    \item \textbf{Resilient Multi-Agent Execution Infrastructure:} We implement a fault-tolerant multi-agent architecture that ensures reliability. It intelligently rotates between different LLMs, retries failed tasks, and combines specialized models to overcome individual API errors or performance issues, guaranteeing a robust generation process (cf. ~\cref{subsec:Resilient Execution}).
    \item \textbf{Open-Source Prototype and Evaluation Metric:} We release an open-source implementation of our system, supporting a wide range of DSPS frameworks and reducing pipeline development time by up to 6.3$\times$ in experimental evaluations (cf.~\cref{sec:performance}). Additionally, we introduce the \textit{Error-Free Score (EFS)}, a novel metric for quantitatively assessing the correctness and completeness of LLM-generated pipelines, offering a more rigorous and holistic evaluation framework than existing measures (cf.~\cref{subsec:results}).
\end{itemize}

\begin{table}
\centering
\setlength{\extrarowheight}{0pt}
\setlength{\aboverulesep}{0pt}
\setlength{\belowrulesep}{0pt}
\caption{Comparison among widely used benchmark suites for data stream processing pipelines (continuous queries). \textdagger~means a framework that can generate infinite benchmark pipelines}
\label{tab:comparison}
\resizebox{\linewidth}{!}{%
\LARGE
\begin{tabular}{cccccc} 
\toprule
\rowcolor[rgb]{0.753,0.753,0.753} \textbf{Benchmark Suite} & \begin{tabular}[c]{@{}>{\cellcolor[rgb]{0.753,0.753,0.753}}c@{}}\textbf{Real-world }\\\textbf{App.}\end{tabular} & \begin{tabular}[c]{@{}>{\cellcolor[rgb]{0.753,0.753,0.753}}c@{}}\textbf{Synthetic }\\\textbf{App.}\end{tabular} & \textbf{DSPSs} & \begin{tabular}[c]{@{}>{\cellcolor[rgb]{0.753,0.753,0.753}}c@{}}\textbf{Unified}\\\textbf{~API}\end{tabular} & \begin{tabular}[c]{@{}>{\cellcolor[rgb]{0.753,0.753,0.753}}c@{}}\textbf{Workload }\\\textbf{Charact.}\end{tabular} \\ 
\midrule
\begin{tabular}[c]{@{}c@{}}Linear Road \\Benchmark \cite{arasu2004linear}\end{tabular} & 1 & - & Aurora & No & No \\ 
\cmidrule(lr){1-6}
\begin{tabular}[c]{@{}c@{}}Yahoo Streaming\\~Benchmark \cite{chintapalli2016benchmarking}\end{tabular} & 1 & - & \begin{tabular}[c]{@{}c@{}}Storm, Flink, \\Spark Streaming\end{tabular} & No & No \\ 
\cmidrule(lr){1-6}
BigDataBench \cite{wang2014bigdatabench} & - & 1 & Spark Streaming & No & No \\ 
\cmidrule(lr){1-6}
StreamBench \cite{lu2014stream} & - & 7 & \begin{tabular}[c]{@{}c@{}}Storm, \\Spark Streaming\end{tabular} & No & Yes \\ 
\cmidrule(lr){1-6}
RIoTBench \cite{shukla2017riotbench} & 4 & - & Storm & Yes & Yes \\ 
\cmidrule(lr){1-6}
HiBench \cite{huang2010hibench} & - & 7 & \begin{tabular}[c]{@{}c@{}}Storm, Flink, \\Spark Streaming\end{tabular} & No & No \\ 
\cmidrule(lr){1-6}
DSPBench \cite{bordin2020dspbench} & 13 & 2 & \begin{tabular}[c]{@{}c@{}}Storm, \\Spark Streaming\end{tabular} & Yes & Yes \\ 
\hline\hline
\rowcolor[rgb]{0.827,0.945,0.875} \textbf{\textsc{AutoStreamPipe}}\textdagger & 7 & 1 & \begin{tabular}[c]{@{}c@{}}Storm, \\Spark, Flink\end{tabular} & Yes & Yes \\
\bottomrule
\end{tabular}
}

\end{table}

\textbf{Paper organization.}
The remainder of this paper is organized as follows: Section~\ref{sec:background} explains background concepts in SP. Section~\ref{sec:architecture} describes the \textsc{AutoStreamPipe} architecture in detail. Section~\ref{sec:performance} presents our comprehensive evaluation of \textsc{AutoStreamPipe}. Section~\ref{sec:related_work} reviews related work. Section~\ref{sec:conclusion} concludes the paper.

%%%%%%%%%%%%%%%%%%%%%%%%%%%%%%%%%%%%%%%%%%%%%%%%%%%%%%%%%%%%%%
\section{Related Work} \label{sec:related_work}

The increasing complexity of DSP pipelines has driven significant research and development efforts toward automating their design and optimization \cite{1639353,yang2023ontology}. These efforts aim to reduce the manual effort required for pipeline creation, improve performance, and make SP more accessible to non-experts. Machine learning (ML)-based tools, for example, employ reinforcement learning, supervised learning, and neural architecture search to generate workflows or ML pipelines similar to data pipelines and query execution \cite{coleman2021wfchef,coleman2023automated,donca2022method,haynes2016pipegen}. However, these techniques typically demand extensive training data and computational power.

Low-code and no-code platforms \cite{lugmayr2023starmap} such as Apache NiFi \cite{10197731,photonics10020210,liu2018generic}, StreamSets \cite{streamsets}, and Node-RED \cite{nodered} offer graphical interfaces that simplify pipeline creation but may lack the flexibility and performance needed for complex or high-throughput scenarios. Rule-based systems, found in many query optimizers within engines like Apache Flink and Spark Streaming, as well as in tools like StreamMine3G \cite{6687445}, utilize predefined heuristics to enhance efficiency but struggle with ambiguous requirements. Template-based solutions offer reusable patterns through cloud services such as Google Cloud Dataflow \cite{ranganathan2023ingestion} and AWS Kinesis, as well as open-source libraries like Streamdrill and Siddhi, although they often fall short in addressing unique needs. Finally, hybrid approaches combine declarative programming models like Flink SQL or KSQL with AI-augmented design tools, offering a balance between ease of use and powerful optimization while requiring advanced infrastructure and expertise.

Recent advances in LLMs have demonstrated remarkable capabilities in automated code generation and software engineering tasks \cite{dong2025survey,wang2023review,nejjar2025llms}. Foundation models such as CodeLlama \cite{roziere2023code}, StarCoder \cite{li2023starcoder}, and CodeGen \cite{nijkamp2022codegen} have been specifically trained on massive code corpora, achieving strong performance across multiple programming languages and paradigms. Transformer-based approaches, such as AlphaCode \cite{li2022competition}, have demonstrated competitive performance in programming competitions. In contrast, CodeT5 \cite{wang2021codet5} utilizes encoder-decoder architectures for code understanding and generation tasks. Commercial tools, such as GitHub Copilot and Amazon CodeWhisperer, have introduced AI-assisted coding into mainstream development workflows, demonstrating the practical viability of LLM-based code generation.
However, these general-purpose code generation models face significant challenges when applied to domain-specific contexts, such as stream processing pipelines. They do not adequately consider streaming-specific constraints, such as stateful operations, windowing semantics, checkpointing requirements, and fault-tolerance guarantees. As a result, the generated code often suffers from incorrect state handling, data loss, or inconsistent event ordering, making it unreliable for real-time execution in DSP environments. Moreover, they typically generate isolated code snippets rather than complete, production-ready pipeline architectures with proper configuration, deployment specifications, and operational considerations. \textsc{AutoStreamPipe} addresses these limitations by combining LLM capabilities with domain-specific reasoning frameworks (HGoT), retrieval-augmented generation for streaming domain knowledge, and multi-agent collaboration to ensure coherent and executable pipeline solutions that satisfy the complex interdependencies inherent in DSPS.

On the other hand, there are some papers published recently that use LLMs to generate workflows for serverless computing \cite{esashi2024action,li2024autoflow,xu2024llm4workflow,wang2025llm4faas}. These papers generate YAML files and are different from our purpose.

\textbf{Summary.} 
Automated SP pipeline design has evolved significantly through several complementary approaches, including machine learning optimization, low-code/no-code platforms, rule-based systems, template-based solutions, and, more recently, LLM-based code generation. Each approach has its own benefits. ML methods are effective for performance optimization, but they require a substantial amount of training data. Low-code platforms make things easier to access, but lose some flexibility. Rule-based systems offer clear optimization but struggle with unclear requirements. Template-based solutions enable quick deployment but often lack customization options. General-purpose LLM code generators demonstrate impressive capabilities but lack understanding of specific streaming requirements, such as stateful operations, windowing rules, checkpointing intervals, and fault tolerance guarantees.
Despite these advancements, current methods often fail to address the various challenges that modern SP systems encounter fully. There has been no effort to combine natural language understanding, domain-specific reasoning for complex streaming connections, multi-agent teamwork for reliable execution, and complete generation of production-ready pipelines with all necessary configuration and deployment details. This gap drives the need for LLM-assisted approaches that mix domain knowledge, structured reasoning frameworks, and strong execution strategies. This integration aims to connect human intent with machine execution, enabling the generation of smart, fully automated pipelines that meet both functional needs and operational constraints.
%%%%%%%%%%%%%%%%%%%%%%%%%%%%%%%%%%%%%%%%%%%%%%%%%%%%%%%%%%%%%%
\section{Background} \label{sec:background}

Over the years, many DSPS have emerged to handle continuous data flows in real-time \cite{fragkoulis2024survey,zubarouglu2021data,cardellini2022runtime,akidau2015dataflow}, including Apache Flink \cite{flink}, Storm \cite{storm}, Spark Streaming \cite{spark}, and Kafka Streaming \cite{kafka}. 
Below, we provide an overview of key components and stages of a typical DSP.

\subsection{Data Stream Pipeline (DSP)} \label{subsec: Data Stream Pipeline}

\begin{figure}[t]
    \centering
    \includegraphics[width=0.95\linewidth]{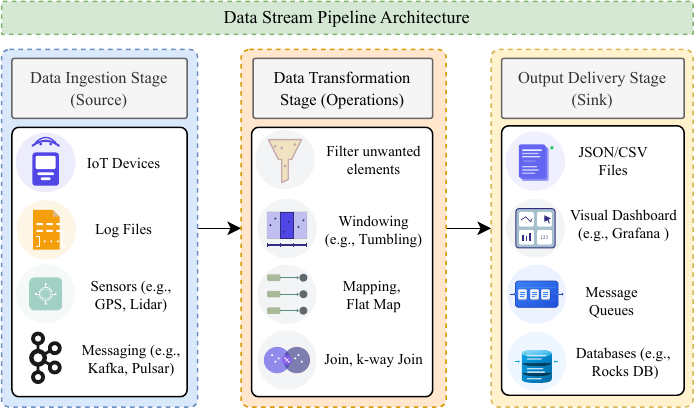}
    \caption{An overview of a data stream pipeline structure.} 
    \label{fig:pipeline_structure}
\end{figure}

\textbf{Definition.} A DSP is an end-to-end system designed to process continuous data streams in real-time (cf. Figure \ref{fig:pipeline_structure}) \cite{cardellini2022runtime}. It consists of interconnected operators that perform tasks such as data ingestion, transformation, and output delivery \cite{cardellini2022runtime,fragkoulis2024survey,zubarouglu2021data}.

Operators are the fundamental building blocks of DSPs, enabling modular, scalable, and flexible data processing. By chaining operators, DSPs move data seamlessly from sources (e.g., sensors, logs) to destinations (e.g., databases, APIs) while applying necessary computations and transformations.

\textbf{Data Ingestion.} Source operators capture raw data from sources such as Kafka topics, files, or sensors and feed it into the pipeline. These operators serve as the starting point for all downstream processing.

\textbf{Data Transformation. } This core stage involves cleaning, filtering, aggregating, and enriching data to make it actionable. Transformations can be stateless (e.g., filtering, mapping) or stateful (e.g., windowing, joins) \cite{carbone2017state}.

\textbf{Output Delivery. } Sink operators store or deliver processed data to external systems such as databases, file systems, or dashboards, ensuring the results are accessible and actionable.

Despite their strengths, DSPSs require significant manual effort to design, configure, and optimize pipelines. This challenge motivates our exploration of LLM-assisted approaches to streamline pipeline generation and enhance usability.

%%%%%%%%%%%%%%%%%%%%%%%%%%%%%%%%%%%%%%%%%%%%%%%%%%%%%%%%%%%%%%
\section{AutoStreamPipe Architecture} \label{sec:architecture}
\begin{figure*}[h]
    \centering
    \includegraphics[width=1\linewidth]{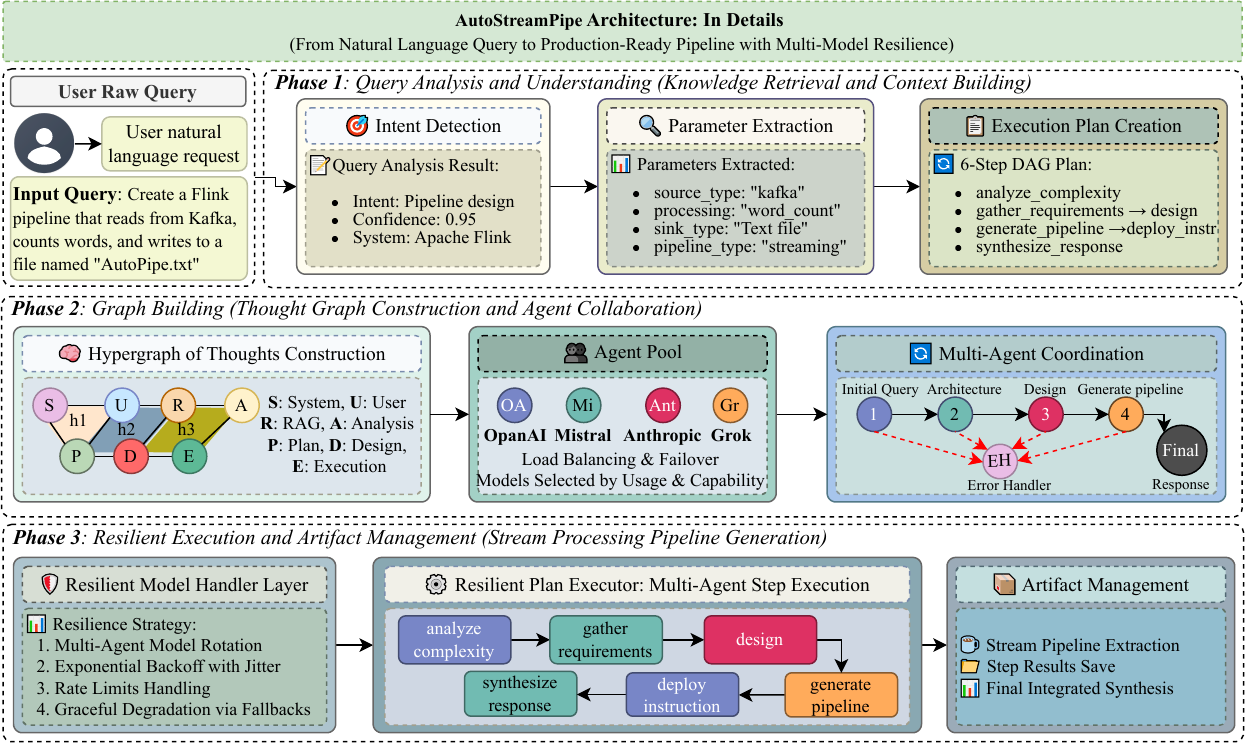}
    \caption{The \textsc{AutoStreamPipe} architecture presents a novel approach to generating production-ready SP pipelines from natural language queries. The system employs a three-phase methodology: (1) query analysis and parameter extraction, (2) hypergraph-based reasoning with multi-agent collaboration, and (3) resilient execution with comprehensive artifact management. This framework harnesses the capabilities of multiple LLM providers while ensuring resilience against API limitations through retry mechanisms and model rotation strategies. }
    \label{fig:AutoStreamPipe-details}
\end{figure*}

In this section, we introduce \textsc{AutoStreamPipe}, a framework that automatically translates natural language queries into SP pipelines. The system utilizes multiple LLMs to ensure robustness against common API limitations such as rate limits, outages, or model unavailability. 
To illustrate how \textsc{AutoStreamPipe} works, we use a running example introduced in box ~\ref{box:wordcount}. Each phase of the pipeline generation process is explained by showing how the system transforms this example step by step, from initial query to executable SP pipeline. The overall architecture is depicted in Figure~\ref{fig:AutoStreamPipe-details}, with detailed algorithms provided in~\cref{alg:AutoStreamPipe,alg:ExecuteStepWithRetry,alg:HGOT,alg:Hyeredgecons}. 

\subsection{Preprocessing Stage}
The system begins by configuring key parameters, such as maximum file size, chunk size, supported file extensions, and SPE settings (Algorithm~\ref{alg:AutoStreamPipe}, line 2). These settings ensure efficient resource utilization and allow for customization without requiring code modifications. This entire stage executes once during system initialization, thereby creating a stable foundation for subsequent operations. This approach addresses the primary limitation of static benchmarks, as outlined in Section~\ref{sec:introduction}, which was their inability to keep pace with rapidly evolving SPE frameworks. AutoStreamPipe implements a dynamic repository management system.

Next, the system identifies and clones target GitHub repositories containing the latest pipeline examples and documentation. By default, it targets Apache Flink, Storm, and Spark, though users can specify alternative frameworks or specific versions via a configuration file. This dynamic approach ensures that AutoStreamPipe always has access to current best practices, emerging patterns, and updated APIs from official SPE repositories. The repository cloning mechanism implements robust fault-tolerance strategies to ensure reliable acquisition of remote resources. When network issues or rate limiting occur, the system employs \textit{exponential backoff with jitter}, a proven resilience technique where retry delays increase exponentially (e.g., 1s, 2s, 4s, 8s) while incorporating randomization to prevent synchronized retry storms across multiple instances. This approach, formalized as $\text{delay} = \text{baseDelay} \times 2^{\text{retries}} \times (0.5 + \text{random()})$ in Algorithm~\ref{alg:ExecuteStepWithRetry} (lines~\ref{alg:es5}-\ref{alg:es6}), significantly enhances system reliability by gracefully handling transient failures without overwhelming external services. 
Once repositories are cloned, they are organized into a structured local hierarchy for easy access and management. If RAG is enabled, the system scans for pipeline documentation, identifies directories with pipeline templates, and indexes components such as source (SO), operator (OP), and sink (SI) (Algorithm~\ref{alg:AutoStreamPipe}). A recursive search efficiently locates directories containing example pipelines, such as the Word Count pipeline.
\begin{algorithm}[t]
\scriptsize
\SetAlgoLined
\caption{Execute Step With Retry}
\label{alg:ExecuteStepWithRetry}
\SetAlgoNoEnd
\SetKwInOut{Input}{Input}
\SetKwInOut{Output}{Output}
\SetKwFunction{FExec}{ESR}
\SetKwProg{Fn}{Function}{:}{}
% Define try-catch blocks
\SetKwBlock{Try}{\textbf{try:}}{\textbf{end try}}
\SetKwBlock{Catch}{\textbf{catch:}}{\textbf{end catch}}
\label{alg:ESR}
\Input{executor, step, plan}
\Output{Result of the executed step}
\Fn{\FExec{executor, step, plan}}{
    retries, maxRetries $\gets$ 0,5\; \label{alg:es1}
    
    \While{retries $<$ maxRetries}{
        \Try{
            result $\gets$ \textsc{ExecuteStep}(step.action, plan.query, step.dependencies)\; \label{alg:es3}
            \Return{result}\; \label{alg:es4}
        }
        \Catch{APIError e}{
            \If{e.isRateLimit}{
                delay $\gets$ baseDelay $\times 2^{\text{retries}} \times (0.5 + \text{random()})$\; \label{alg:es5}
                \textsc{Sleep}(delay)\; \label{alg:es6}
            }
            \ElseIf{e.isQuotaExceeded}{
                \textsc{SwitchToNextModel}(executor)\; \label{alg:es7}
            }
        }
        retries $\gets$ retries + 1\; \label{alg:es8}
    }
    
    \Return{\textsc{GenerateFallbackResult}(step)}\; \label{alg:es9}
}
\end{algorithm}
Finally, robust error handling ensures that issues with individual files don’t disrupt the workflow. The output includes annotated code examples with metadata and SHA-256 checksums, balancing readability for humans with machine processability while preserving important context.

\subsubsection{Error Handling and Output Preparation}

Robust error handling is integral to this phase, ensuring that issues with individual files do not disrupt the overall workflow. The system implements graceful degradation strategies that log and skip corrupted or inaccessible files rather than terminating the process. File integrity is verified through SHA-256 checksums, which detect any corruption during transfer or storage. When processing errors occur, the system continues with available resources while maintaining a comprehensive error log for diagnostic purposes.

The output of this phase includes annotated code examples enriched with metadata such as SPE version, component type, and usage context. This dual representation balances human readability with machine processability, preserving important contextual information that guides subsequent pipeline generation steps. By establishing this comprehensive, dynamically updated knowledge base, Phase 1 creates the foundation upon which all subsequent AutoPipe operations build, effectively solving the benchmark obsolescence problem through continuous integration of evolving SPE documentation.

\begin{algorithm}[t]
\scriptsize
\SetAlgoLined
\caption{AutoStreamPipe}
\SetAlgoNoEnd
\SetKwInOut{Input}{Input}
\SetKwInOut{Output}{Output}
\label{alg:AutoStreamPipe}
\Input{User query, Streaming system selection, Config options}
\Output{Stream processing pipeline solution with code and documentation}
\SetKwFunction{FAuto}{AutoStreamPipe}
\SetKwProg{Fn}{Function}{:}{}
\Fn{\FAuto{query, streamingSystem, options}}{ \label{alg:ASP1}
    models $\gets$ \textsc{InitializeModels}(options.modelsList, options.backupModels)\;  \label{alg:ASP3}
    planningModel $\gets$ \textsc{GetPlanningModel}(models[0])\; \label{alg:ASP3}
    retryHandler $\gets$ \textsc{IRH}(models)\; \label{alg:ASP4}
    
    % \tcp{Phase 1: Query Understanding}
    intent $\gets$ \textsc{DetectIntentWithRetry}(retryHandler, query)\; \label{alg:ASP5}
    parameters $\gets$ \textsc{EPR}(retryHandler, query, intent)\; \label{alg:ASP6}
    executionPlan $\gets$ \textsc{CEP}(query, intent, streamingSystem, options.useRAG)\; \label{alg:ASP7}
    
    % \tcp{Phase 2: Graph Building}
    thoughtGraph $\gets$ \textsc{HGoT Construction}(streamingSystem, query, options)\; \label{alg:ASP8}
    \If{options.useRAG}{\label{alg:ASP9}
        ragDocs $\gets$ \textsc{RRD}(query, streamingSystem)\; \label{alg:ASP10}
        \ForEach{doc in ragDocs}{\label{alg:ASP11}
            \textsc{AddNodeToGraph}(thoughtGraph, doc, "rag")\;\label{alg:ASP12}
        }
    }
    \textsc{NodeToGraph}(thoughtGraph, "QueryAnalysisResult", "analysis")\; \label{alg:ASP13}
    \textsc{NodeToGraph}(thoughtGraph, "ExecutionPlan", "plan")\; \label{alg:ASP14}
    
    % \tcp{Phase 3: Multi-Agent Collaboration \& Resilient Execution}
    planExecutor $\gets$ \textsc{CreateResilientExecutor}(models, thoughtGraph, streamingSystem)\; \label{alg:ASP15}
    \ForEach{step in executionPlan.steps}{\label{alg:ASP16}
        \While{step has unmet dependencies}{\label{alg:ASP17}
            \textbf{continue}\;\label{alg:ASP18}
        }
        result $\gets$ \textsc{ESR}(planExecutor, step, executionPlan)\; \label{alg:ASP19}
        \textsc{SaveStepResult}(result, step)\; \label{alg:ASP20}
        \If{result contains JavaCode}{ \label{alg:ASP21}
            javaFile $\gets$ \textsc{ExtractAndSaveJavaCode}(result)\; \label{alg:ASP22}
        }
        \textsc{MarkStepCompleted}(step, result)\; \label{alg:ASP23}
    }
    
    % \tcp{Phase 4: Artifact Management}
    finalResponse $\gets$ \textsc{SynthesizeResponse}(executionPlan, thoughtGraph)\; \label{alg:ASP24}
    summary $\gets$ \textsc{CreateSessionSummary}(query, intent, streaming system)\; \label{alg:ASP25}
    \textsc{SaveMemory}(query, finalResponse, thoughtGraph)\; \label{alg:ASP26}
    \Return{(finalResponse, summary)}\;\label{alg:ASP27}
}
\end{algorithm}

\subsection{Phase 1: Query Analysis and Understanding}\label{sec:Query_analysis}

The first phase of \textsc{AutoStreamPipe} processes natural language inputs through a series of analytical steps, as outlined in Algorithm~\ref{alg:AutoStreamPipe}.

\textbf{Step 1: Query Analysis.}
Intent Detection begins by interpreting the user’s request through intent understanding, a two-stage process combining efficiency and adaptability. First, it applies fast regular expression pattern matching (e.g., detecting phrases like in the box ~\ref{box:wordcount} to infer an \texttt{intent type} of \texttt{pipeline design}). For ambiguous or complex queries, an LLM-powered intent detector is reconstructed, which structures the input into a lightweight \texttt{QueryIntent} object containing: \textbf{(1)} a high-level category (e.g., \texttt{Pipeline design}), \textbf{(2)} a confidence score (a float between 0 and 1), and (3) extracted parameters (e.g., framework = "Apache Flink", \texttt{window\_size} = 5 min).

\textbf{Step 2: Parameter Extraction.}
This builds upon the detected intent to identify specific requirements embedded in the query. Based on the intent type (Algorithm~\ref{alg:AutoStreamPipe}, line 9), the system employs specialized extraction using an LLM that targets parameters relevant to that particular intent. For pipeline design intents, these parameters include the data source type ("kafka" in the example), processing operations ("word\_count"), sink type ("Text file"), and pipeline type ("streaming"). The parameter extraction process employs similar resilient retry capabilities to intent detection, ensuring robustness when interfacing with external LLM providers. The extraction process balances precision with flexibility, identifying explicit parameters while inferring implicit ones based on domain knowledge of SP systems. 
The resulting parameter set formalizes the user's requirements, transforming ambiguous natural language into structured data to drive pipeline generation.

\textbf{Step 3: Create Execution Plan.}
This step concludes phase 2 by constructing a directed acyclic graph (DAG) that outlines the sequential steps needed to generate the requested pipeline (Algorithm~\ref{alg:AutoStreamPipe}, line 10). 
% \end{lstlisting}
\begin{table*}[t]
\centering
\setlength{\extrarowheight}{0pt}
% \addtolength{\extrarowheight}{\aboverulesep}
% \addtolength{\extrarowheight}{\belowrulesep}
\setlength{\aboverulesep}{0pt}
\setlength{\belowrulesep}{0pt}
\caption{Comparison of Advanced Reasoning Frameworks}
\label{tab:LLMReasoningcomparison}
\resizebox{\linewidth}{!}{%
\begin{tabular}{lcccccc} 
\toprule
\rowcolor[rgb]{0.816,0.816,0.816} \textbf{Feature} & \textbf{Chain of Thought} & \textbf{CoT-SC} & \textbf{Tree of Thoughts} & \textbf{Layer of Thoughts} & \textbf{Graph of Thoughts} & \textbf{Hypergraph of Thoughts} \\ 
\hline
\textbf{Structure} & Linear sequence & Parallel linear sequences & Hierarchical tree & Layered graph (DAG) & Directed graph & Hypergraph \\
\textbf{Node Connections} & \begin{tabular}[c]{@{}c@{}}One-to-one \\(sequential)\end{tabular} & \begin{tabular}[c]{@{}c@{}}One-to-one \\(within chains)\end{tabular} & \begin{tabular}[c]{@{}c@{}}One-to-many\\~(branching)\end{tabular} & \begin{tabular}[c]{@{}c@{}}Many-to-many\\~(layer-to-layer)\end{tabular} & \begin{tabular}[c]{@{}c@{}}Many-to-many \\(pairwise)\end{tabular} & \begin{tabular}[c]{@{}c@{}}Many-to-many \\(group-based)\end{tabular} \\
\textbf{Reasoning Flow} & Unidirectional & Parallel unidirectional & Hierarchical & Layered, feed-forward & Networked & Higher-order networked \\
\textbf{Backtracking} & Limited/None & None (selection by voting) & Natural & via refinement layers & Supported & ~group backtracking \\
\textbf{Parallelism} & Limited & Very high & Moderate & High (within layers) & High & Very high \\
\textbf{Multi-constraint} & Poor & Moderate & Moderate & Good & Good & Excellent \\
\textbf{Group Inference} & Not supported & Limited (via aggregation) & Limited & Supported & Limited & Native support \\
\textbf{Complexity} & $O(n)$ & $O(k \cdot n)$ & $O(b^d)$ & Layer-dependent & $O(n^2)$ & $O(2^n)$ worst case \\
\textbf{Suitable Problems} & Sequential reasoning & Robustness via diversity & Hierarchical decomposition & Layered refinement & Networked dependencies & Complex constraint satisfaction \\
\bottomrule
\end{tabular}
}
\footnotesize \textit{Note:}
$n$: number of reasoning steps/nodes,
$k$: number of chains in CoT-SC,
$b$: branching factor in ToT,
$d$: depth of the tree in ToT
\end{table*}

Based on the identified intent and extracted parameters, the system customizes the plan with steps suited for the request type, typically including: \texttt{analyze complexity} to assess the pipeline's computational characteristics, \texttt{gather requirements} to formalize specifications, design to develop the architectural structure, \texttt{generate pipeline} to create implementation code, \texttt{deploy instructions} to provide operational guidance, and \texttt{synthesize response} to produce the final output. The plan establishes dependencies between steps to ensure a coherent execution where each stage builds on its predecessor, offering a clear roadmap for progress. The execution plan converts user intent into a concrete procedure for generating the desired pipeline.

% \begin{infobox}
% \small
% \caption{\small Word Count Pipeline (Complete Example Version)}
% \label{box:wordcount}
% \begin{mybox}
% Create an Apache Flink streaming application that processes text data with the following specifications:
% \begin{itemize}[wide]
%   \item \textbf{Source:} Kafka topic ``input-text'' (bootstrap servers: \texttt{localhost:9092}, consumer group: \texttt{word-count-group})
%   \item \textbf{Input format:} Plain text messages with UTF-8 encoding
%   \item \textbf{Processing:} Split messages by whitespace regex \verb!"\s+"!, convert to lowercase, filter words with length $\ge 3$
%   \item \textbf{Windowing:} 30-second tumbling windows for aggregation
%   \item \textbf{Output:} Local file system at \texttt{/output/word-counts.txt} with format ``word,count,timestamp''
%   \item \textbf{Parallelism:} 4 for source, 8 for processing, 2 for sink
%   \item \textbf{Checkpointing:} Every 10 seconds with sqlite3 state backend
%   \item \textbf{Error handling:} Dead letter queue for malformed messages to Kafka topic ``dlq-text''
% \end{itemize}
% \end{mybox}
% \end{infobox}

\begin{infobox}
\footnotesize
\caption{Word Count Pipeline (Complete Example Version)}
\label{box:wordcount}
\begin{mybox}[frametitle={Word Count Pipeline (Complete Example Version)}]
Create an Apache Flink streaming application that processes text data with the following specifications:
\begin{itemize}[wide]
  \item \textbf{Source:} Kafka topic ``input-text'' (bootstrap servers: \texttt{localhost:9092}, consumer group: \texttt{word-count-group})
  \item \textbf{Input format:} Plain text messages with UTF-8 encoding
  \item \textbf{Processing:} Split messages by whitespace regex \verb!"\s+"!, convert to lowercase, and filter words with length $\ge 3$
  \item \textbf{Windowing:} 30-second tumbling windows for aggregation
  \item \textbf{Output:} Local file system at \texttt{/output/word-counts.txt} with format ``word,count,timestamp''
  \item \textbf{Parallelism:} 4 for source, 8 for processing, 2 for sink
  \item \textbf{Checkpointing:} Every 10 seconds with sqlite3 state backend
  \item \textbf{Error handling:} Dead letter queue for malformed messages to Kafka topic ``dlq-text''
\end{itemize}
\end{mybox}
\end{infobox}

The second phase of \textsc{AutoStreamPipe} converts unstructured natural language into structured representations to guide the generation of the pipeline. The process begins when the user submits a request as raw text input to Algorithm~\ref{alg:AutoStreamPipe}, as shown in the box \ref{box:wordcount}. The system applies preliminary preprocessing, normalizes the text, removes extra whitespace, standardizes punctuation, and recognizes key entities to formalize implicit requirements and structure the pipeline design. 

\subsection{Phase 2: Graph Building and Agent Collaboration} \label{subsec: Graph Building}
The second phase establishes the reasoning framework and agent coordination infrastructure that enable the generation of sophisticated pipelines. In the context of automated pipeline generation, \textit{reasoning} refers to the systematic process of deriving conclusions, making design decisions, and solving problems through logical inference and the integration of knowledge. Specifically, AutoStreamPipe must reason about interdependent components: determining compatible data sources, selecting appropriate operators, configuring state management, and ensuring fault tolerance. All while maintaining consistency across these interconnected decisions.

Pipeline synthesis for streaming data platforms (\emph{e.g.}, Apache Flink) involves interdependent design decisions, including data source configuration, operator chaining, state management, and fault tolerance. For instance, choosing a Kafka source influences the selection of serialization format, which in turn affects the implementation of operators, which in turn constrains sink configuration. These multi-way dependencies create a reasoning challenge: a decision in one component can simultaneously invalidate or necessitate changes in multiple other components.  Pairwise graph structures for reasoning like CoT, GoT, where edges connect only two nodes at a time, struggle to encode such multifactor constraints. They force the system into repeated backtracking when dependencies are violated, leading to inconsistent partial plans and computational inefficiency.  By contrast, HGoT's hyperedges enable the system to bind \emph{sets} of related decisions into coherent reasoning units, facilitating global consistency checks and synchronized updates across multiple dependent components.

\begin{figure}
    \centering
    \includegraphics[width=1\linewidth]{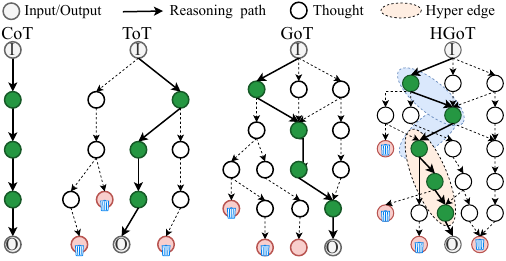}
    \caption{Illustration of the CoT, ToT, GoT, and HGoT reasoning process pipeline. This structure enables higher-order dependencies, dynamic pruning of infeasible paths (represented by red trash-bin icons), and adaptive traversal toward an optimal solution. }
    \label{fig:reasoning}
\end{figure}

\textbf{Step 1: Hypergraph of Thoughts Construction.}
The reasoning framework at the heart of \textsc{AutoStreamPipe} employs a novel cognitive architecture known as the \textit{Hypergraph of Thoughts (HGoT)}. To understand HGoT's advantages, we first examine the evolution of reasoning frameworks in LLMs (see \cref{tab:LLMReasoningcomparison}). Recent reasoning frameworks, such as Chain of Thought (CoT) \cite{wei2022chain}, Multiple Chains of Thought (CoT-SC) \cite{yoran2023answering}, Tree of Thoughts (ToT) \cite{yao2023tree}, Graph of Thoughts (GoT) \cite{besta2024graph}, and Layer of Thoughts (LoT) \cite{fungwacharakorn2024layer}, incrementally improve flexibility, complexity, and interpretability. However, these approaches share a fundamental limitation: they represent reasoning structures using pairwise connections among nodes or layers, as illustrated in Table~\ref{tab:LLMReasoningcomparison} and Figure~\ref{fig:reasoning}. This pairwise constraint creates a representational bottleneck when modeling real-world problems with multi-way dependencies. Consider pipeline state management: the choice of state backend (e.g., RocksDB vs. heap-based) simultaneously affects checkpointing strategy, recovery time objectives, memory configuration, and operator parallelism. Representing this four-way dependency using pairwise edges requires creating multiple redundant connections and complex coordination logic, obscuring the fundamental unity of these interconnected decisions.

To overcome these limitations, \textsc{AutoStreamPipe} implements the HGoT architecture, which extends traditional graphs by introducing hyperedges that can simultaneously connect multiple nodes. This hypergraph structure offers a more expressive representation of complex, multifactor constraints and relationships in practical applications, such as scheduling with interdependent tasks.

In the Hypergraph of Thoughts, the nodes represent individual reasoning steps, hypotheses, or partial solutions. Hyperedges link groups of nodes to capture multi-way relationships and interdependencies directly. This structure enables higher-order reasoning and synchronized iterative refinement, ensuring consistency across interconnected reasoning elements. The system iteratively evaluates hyperedge constraints, simultaneously adjusting the involved nodes by refining solutions, merging compatible ideas, or pruning infeasible options.
HGoT improves \textsc{AutoStreamPipe}'s ability to manage complexity and converge toward globally coherent solutions by enabling synchronized updates and capturing intricate interdependencies among reasoning steps. This capability supports flexible answer aggregation, accommodating single/multiple outcomes depending on task requirements.

\paragraph{Formal definition}
We model reasoning in \textsc{AutoStreamPipe} as an HGoT, formally defined by the tuple
$H=(V,E,\Phi,\Psi)$, where: 
\begin{itemize}[leftmargin=*]
  \item $V=\{v_{1},\dots,v_{n}\}$ is the set of thought vertices, each encoding an individual cognitive unit.
  \item $E=\{e_{1},\dots,e_{m}\}$ is the set of hyperedges, with $e_{j}\subseteq V$ able to link an arbitrary non‑empty subset of vertices and thereby capture higher‑order relations.
  \item $\Phi:V\rightarrow\mathcal V$ assigns each vertex a semantic embedding in space $\mathcal V$.
  \item $\Psi:E\rightarrow\mathcal E$ maps each hyperedge to an embedding in space $\mathcal E$ that reflects the semantics of the underlying relation.
\end{itemize}
\paragraph{Thought structure}
Each vertex $v_{i}$ is a quadruple $v_{i}=(c_{i},\tau_{i},\sigma_{i},\mathbf x_{i})$ consisting of \emph{content} $c_{i}$, \emph{type} $\tau_{i}$ (premise, hypothesis, \emph{etc.}), \emph{confidence} $\sigma_{i}\!\in\![0,1]$ and embedding $\mathbf x_{i}=\mathcal L(c_{i})$ obtained from a language model encoder $\mathcal L$. The confidence score is computed as:
\begin{equation}
\sigma_i = f_{\text{conf}}(v_i;\mathcal C,H_t)
        = \alpha\,\mathrm{rel}(v_i,\mathcal C)
        + \beta\,\mathrm{cons}(v_i,H_t)
        + \gamma\,\mathrm{spec}(v_i),
\end{equation}
where $\alpha + \beta + \gamma = 1$ and $H_t$ is the hypergraph state at time $t$.

\paragraph{Hyperedge semantics}
Each hyperedge $e_{j}$ is annotated as $e_{j}=(S_{j},T_{j},r_{j},w_{j},t_{j})$ where $S_{j}$ and $T_{j}$ are source and target vertex subsets respectively. $r_{j}$ is a relation label (e.g.\ \emph{causation, refinement}), $w_{j}\!\in\!\mathbb R$ is a weight, and $t_{j}$ is a temporal stamp.  Directionality is indicated when $S_{j}\neq T_{j}$. The hyperedge weight is computed as:
\begin{equation}
w_j = \frac{1}{|S_j| \cdot |T_j|} \sum_{v_i \in S_j} \sum_{v_k \in T_j} \text{sim}(\mathbf{x}_i, \mathbf{x}_k) \cdot \text{relevance}(r_j, \tau_i, \tau_k)
\end{equation}
where $\text{sim}(\mathbf{x}_i, \mathbf{x}_k) = \frac{\mathbf{x}_i \cdot \mathbf{x}_k}{||\mathbf{x}_i|| \cdot ||\mathbf{x}_k||}$ is cosine similarity.

\paragraph{Core reasoning operations}
HGoT exposes five core operators that together support construction and analysis:

\begin{enumerate}[wide]
  \item \textbf{Generate:} Creates a new vertex from context $\mathcal C$ and $k$ prior thoughts (\( \mathcal{C} \times V^k \rightarrow V \)).
  \item \textbf{Connect:}
        Inserts a new hyperedge linking any subset of vertices with a relation label and weight ($2^{V}\times\mathcal R\times\mathbb R\!\to\!E$).
  \item \textbf{Evaluate:}
       Updates confidence scores ($V\times\mathcal C\!\to\![0,1]$).
  \item \textbf{Refine:}
        Outputs an improved version of a thought ($V\times\mathcal C\!\to\!V$).
  \item \textbf{Traverse:}
     Selects the next vertex sequence under strategy $\mathcal S$ which can be confidence-guided or relation-guided ($V\times E\times\mathcal S\!\to\!V^{*}$).
     The traversal strategies are defined as: \emph{confidence-guided} traversal selects $\pi_{\text{conf}}(v_i) = \arg\max_{v_j \in \mathcal{N}(v_i)} \sigma_j$ where $\mathcal{N}(v_i) = \{v_j \in V : \exists e_k \in E \text{ such that } \{v_i, v_j\} \subseteq e_k\}$, \emph{relation-guided} traversal uses $\pi_{\text{rel}}(v_i, r) = \{v_j \in V : \exists e_k \in E \text{ with } r_k = r \text{ and } \{v_i, v_j\} \subseteq e_k\}$; and \emph{multi-objective} traversal applies $\pi_{\text{multi}}(v_i) = \arg\max_{v_j \in \mathcal{N}(v_i)} [\alpha \cdot \sigma_j + \beta \cdot \text{novelty}(v_j) + \gamma \cdot \text{relevance}(v_j, \mathcal{C})]$ where $\text{novelty}(v_j) = 1 - \max_{v_k \in V \setminus \{v_j\}} \text{sim}(\mathbf{x}_j, \mathbf{x}_k)$ and $\text{relevance}(v_j, \mathcal{C}) = \text{sim}(\mathbf{x}_j, \mathcal{L}(\mathcal{C}))$.
\end{enumerate}

The combination of these elements gives the hypergraph with constructive, analytic, and exploratory capabilities that generalize prominent predecessors:
\emph{Chain of Thought}~\cite{wei2022chain}, \emph{Tree of Thoughts}~\cite{yao2023tree}, and \emph{Graph of Thoughts}~\cite{besta2024graph} are all recovered as special cases when $|e_{j}|=2$ with path, tree, or simple‑graph topologies, respectively.
\paragraph{Expressivity and complexity}
By advancing reasoning from binary relationships to hyperrelationships, HGoT demonstrates a significantly enhanced expressive power compared to CoT, ToT, and GoT. Importantly, its space complexity remains at \(O(|V| + \sum_{e_{j} \in E} |e_{j}|)\), which is often more efficient than the \(O(|V|^{2})\) required for enumerating all pairwise relations.

\begin{figure}
    \centering
    \includegraphics[width=1\linewidth]{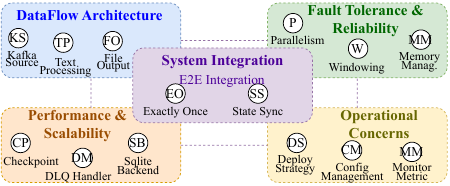}
    \caption{ HGoT applied to Wordcount pipeline design. The diagram illustrates how HGoT simultaneously considers multiple interdependent design dimensions through hyperedges (dashed ellipses): Data Flow Architecture (blue), Performance and Scalability (orange), Reliability and Fault Tolerance (green), and Operational Concerns (yellow). Individual thoughts (circles) within each hyperedge represent specific requirements and constraints. The central System Integration hyperedge (purple) captures cross-cutting concerns, such as exactly-once semantics and state synchronization.
    }
    \label{fig:hgot-flink-pipeline}
\end{figure}

\paragraph{HGOT construction}
The process begins by constructing an initial graph with system and user nodes (Algorithm~\ref{alg:AutoStreamPipe}, line 11), establishing the fundamental context for reasoning. When RAG is enabled, document nodes containing relevant domain knowledge are incorporated into the graph (Algorithm~\ref{alg:AutoStreamPipe}, lines 12-15), creating connections between user requirements and system capabilities. Analysis and plan nodes are then added (Algorithm~\ref{alg:AutoStreamPipe}, lines 16-17), representing the system's evolving understanding of the problem and approach to solving it. The resulting structure forms a hypergraph where edges can connect multiple nodes simultaneously, representing complex relationships between different reasoning elements. 
As visualized in the Figure~\ref{fig:AutoStreamPipe-details}, node types include System ($S$) containing LLM, User ($U$) representing user queries, (retrieved domain knowledge (via RAG ($R$)), Analysis ($A$) containing query interpretation, Plan ($P$) holding execution strategies, Design ($D$) for architectural decisions, and Execution ($E$) tracking implementation progress. These nodes are interconnected through standard edges and hyperedges, creating a rich structure that captures the multifaceted reasoning necessary for pipeline generation. 
Algorithm~\ref{alg:HGOT} begins by creating \emph{system} ($S$) and \emph{user} ($U$) vertices, seeding $V_{0}$ with the dialogue context.  If RAG is active, document vertices $R$ containing domain knowledge are added and connected to $S,U$ via \textbf{Connect}.  Subsequent calls to \textbf{Generate} and \textbf{Refine} populate \emph{analysis} ($A$) and \emph{plan} ($P$)
vertices, progressively elaborating the solution space. The hyperedge construction process, as outlined in Algorithm~\ref{alg:Hyeredgecons}, employs similarity-based clustering to identify coherent thought groups and automatically determines appropriate relation types and weights. These hyperedges naturally encode constraints such as \textit{``the window size must align with checkpointing intervals''} or \textit{``error-handling strategy depends jointly on sink semantics and broker QoS''}. During traversal, Algorithm~\ref{alg:HGOT}\,(lines~12--16) alternates a \emph{confidence‑guided} and \emph{relation‑guided} policy to balance exploitation of promising subgraphs against exploration for novel but relevant alternatives.

\begin{algorithm}[h]
\caption{HGoT Construction}
\label{alg:HGOT}
\SetAlgoLined
\scriptsize
\SetAlgoNoEnd
\DontPrintSemicolon
\KwIn{User requirements $U$, System constraints $S$, Knowledge base $R$}
\KwOut{Optimal pipeline design $D^*$}

$V_0 \leftarrow S \cup U \cup R$\;
$E_0 \leftarrow \text{Conn.}(S \cup U, \text{``context''}) \cup \text{Conn.}(R, S \cup U, \text{``knowledge''})$\;
$H_0 \leftarrow (V_0, E_0, \Phi, \Psi)$\;

\For{$t \leftarrow 1$ \KwTo $\text{max\_iterations}$}{
    $A_t \leftarrow \{\text{Generate}(\text{context}, V_{t-1}) \text{ for analysis aspects}\}$\;
    $V_t \leftarrow V_{t-1} \cup A_t$\;
    $P_t \leftarrow \{\text{Generate}(\text{context} \cup A_t, V_t) \text{ for plan components}\}$\;
    $V_t \leftarrow V_t \cup P_t$\;
    $E_t \leftarrow E_{t-1} \cup \text{Algorithm}~\ref{alg:Hyeredgecons}(V_t)$

    \eIf{$t \bmod 2 = 1$}{
        $\text{next\_vertices} \leftarrow \pi_{\text{conf}}(\text{current\_vertex})$\; %\tcp*{Confidence-guided traversal}
    }{
        $\text{next\_vertices} \leftarrow \pi_{\text{rel}}(\text{current\_vertex}, \text{``dependency''})$\;% \tcp*{Relation-guided traversal}
    }

    \ForEach{$v \in \text{next\_vertices}$}{
        $v_{\text{refined}} \leftarrow \text{Refine}(v, \text{context})$\;
        $V_t \leftarrow V_t \cup \{v_{\text{refined}}\}$\;
    }

    \If{$\text{convergence\_criterion\_met}(H_t)$}{
        \textbf{break}\;
    }
}
$D^* \leftarrow \text{extract\_optimal\_design}(H_t)$\;
\Return{$D^*$}\;
\end{algorithm}

\begin{algorithm}[h]
\caption{Hyperedge Construction}
\label{alg:Hyeredgecons}
\SetAlgoLined
\SetAlgoNoEnd
\footnotesize
Compute pairwise similarities $S_{i,j} = \text{sim}(v_i, v_j)$ for all $v_i, v_j \in V$\;
Apply hierarchical clustering to obtain candidate groups $G$\;
\For{each cluster $g_k \in G$ with $|g_k| \geq 2$}{
  Identify the common theme $\theta_k$ across thoughts in $g_k$\;
  Assign relation type $r_k$ based on $\theta_k$\;
  Compute weight $w_k = \frac{1}{|g_k|} \sum_{v_i, v_j \in g_k} S_{i,j}$\;
  Create hyperedge $e_k = (g_k, r_k, w_k, 0, t)$\;
}
\end{algorithm}

\paragraph{Example} 
Figure~\ref{fig:hgot-flink-pipeline} illustrates the hypergraph decomposition of the Flink pipeline into six critical hyperedges that capture the multi-dimensional relationships between architectural components:
\[\small
\begin{aligned}
e_1 &= (\{KS, TP, FO\}, \emptyset, \text{``data\_flow''}, w_1, 0, t_1) \\
e_2 &= (\{P, W, MM\}, \emptyset, \text{``performance\_optimization''}, w_2, 0, t_2) \\
e_3 &= (\{CP, DM, SB\}, \emptyset, \text{``fault\_tolerance''}, w_3, 0, t_3) \\
e_4 &= (\{MM, CM, DS\}, \emptyset, \text{``operational\_Concern''}, w_4, 0, t_4) \\
e_5 &= (\{P, CP, W\}, \{EO\}, \text{``perf\_reliability\_tradeoff''}, w_5, 1, t_5) \\
e_6 &= (\{KS, CP, SB, DM, EO, SS\}, \{EC\}, \text{``system\_integration''}, w_6, 1, t_6)
\end{aligned}
\]
Hyperedge $e_1$ represents the core data flow architecture, connecting the Kafka Source (KS) through Text Processing (TP) to File Output (FO), forming the primary ingestion and transformation path. Hyperedge $e_2$ captures performance and scalability constraints, where the parallelism configuration $P_{4-8-2}$ must be harmonized with the $30$-second windowing strategy $W_{30s}$ and task memory allocation (MM) to avoid bottlenecks. Hyperedge $e_3$ models the fault-tolerance mechanism, requiring tight coordination between $10$-second checkpoint intervals $CP_{10s}$, Dead Letter Queue (DLQ) handling, and State Backend (SB) persistence to ensure recovery consistency.

The directed hyperedge $e_5$ is critical: it represents the performance–reliability trade-off, where the interplay between parallelism levels, checkpoint frequency, and window duration collectively determines the feasibility of exactly-once (EO) processing semantics. Violation of this constraint leads to state inconsistency or degraded throughput.

Most importantly, hyperedge $e_6$ highlights the main system integration constraint, marked in purple in Figure~\ref{fig:hgot-flink-pipeline}. It connects six key components: Kafka Source (KS), checkpointing (CP$_{10s}$), state backend (SB), DLQ, exactly-once semantics (EO), and snapshotting (SS), with emergent consistency (EC) as the dependent outcome. This hyperedge ensures end-to-end correctness and is vital for reliable and consistent stream processing across various components.

Since these five decisions exist within a single higher-order constraint, any change, such as reducing the window from 30 seconds to 10 seconds, prompts a review of the other four parameters. This aim is to maintain exactly-once semantics through synchronized adjustments. The checkpoint interval must maintain a 1:1 ratio (30 seconds to 10 seconds), parallelism must remain optimal for the new window, and the DLQ policy needs to adjust timeout thresholds. This synchronized adjustment isn’t feasible in pairwise-edge frameworks (CoT, ToT, GoT). It demonstrates how HGoT guides \textsc{AutoStreamPipe} towards a coherent design point $D^{*}$. The process culminates in the executable Flink specification presented in the box ~\ref{box:wordcount}.

\begin{algorithm}[t]
\scriptsize
\SetAlgoLined
\caption{Execute Step With Retry}
\SetAlgoNoEnd
\SetKwInOut{Input}{Input}
\SetKwInOut{Output}{Output}
\SetKwFunction{FExec}{ESR}
\SetKwFunction{FSwitch}{SwitchToNextModel}
\SetKwProg{Fn}{Function}{:}{}
% Define try-catch blocks
\SetKwBlock{Try}{\textbf{try:}}{\textbf{end try}}
\SetKwBlock{Catch}{\textbf{catch:}}{\textbf{end catch}}
\label{alg:ESR}
\Input{executor, step, plan}
\Output{Result of the executed step}

\Fn{\FExec{executor, step, plan}}{
    retries, maxRetries $\gets$ 0, 5\; \label{alg:es1}
    
    \While{retries $<$ maxRetries}{
        \Try{
            result $\gets$ \textsc{ExecuteStep}(step.action, plan.query, step.dependencies)\; \label{alg:es3}
            \Return{result}\; \label{alg:es4}
        }
        \Catch{APIError e}{
            \If{e.isRateLimit}{\label{alg:es5}
                delay $\gets$ baseDelay $\times 2^{\text{retries}} \times (0.5 + \text{random()})$\; \label{alg:es5-1}
                \textsc{Sleep}(delay)\; \label{alg:es6}
            }
            \ElseIf{e.isQuotaExceeded}{
                \textsc{SwitchToNextModel}(executor)\; \label{alg:es7}
            }
        }
        retries $\gets$ retries + 1\; \label{alg:es8}
    }
    
    \Return{\textsc{GenerateFallbackResult}(step)}\; \label{alg:es9}
}
\Fn{\FSwitch{executor}}{
    currentIndex $\gets$ executor.currentModelIndex\; \label{alg:switch1}
    modelPool $\gets$ executor.modelPool\; \label{alg:switch2}
    
    \If{currentIndex $<$ $|$modelPool$|$ $-$ 1}{\label{alg:switch3}
        executor.currentModelIndex $\gets$ currentIndex + 1\; \label{alg:switch4}
        executor.activeModel $\gets$ modelPool[currentIndex + 1]\; \label{alg:switch5}
    }
    \Else{\label{alg:switch6}
        executor.currentModelIndex $\gets$ 0\; \label{alg:switch7}
        executor.activeModel $\gets$ modelPool[0]\; \label{alg:switch8}
    }
}
\end{algorithm}

\textbf{Step 2: Agent Pool Configuration.}
Multiple models from different providers are initialized (Algorithm~\ref{alg:AutoStreamPipe}, lines 4-5), including specialized models from OpenAI, Mistral, Anthropic, and Groq, as depicted in Figure~\ref {fig:AutoStreamPipe-details}. These models are organized into a managed pool with sophisticated load balancing capabilities that distribute tasks based on model strengths, availability, and quota consumption. Failover mechanisms are configured to automatically switch between models when rate limits or other failures are encountered, ensuring continuous operation even when individual providers experience limitations. The agent selection system incorporates a static configuration based on known model capabilities and a dynamic adaptation based on observed performance, creating an evolving selection mechanism that optimizes quality and reliability. This agent pool represents a key innovation in \textsc{AutoStreamPipe}, transcending the limitations of single-model approaches by creating a heterogeneous AI team with complementary capabilities.

\textbf{Step 3: Multi-Agent Coordination Setup. } 
A resilient plan executor is created (Algorithm~\ref{alg:AutoStreamPipe}, line 20) that manages the assignment of tasks to specific models based on their capabilities and availability. The coordination system implements error handling mechanisms, as visualized in Figure~\ref {fig:AutoStreamPipe-details} by red dashed lines connecting to the Error Handler (EH) node. These connections represent the system's ability to detect failures at any point in the process and initiate appropriate recovery actions. Different steps in the pipeline generation process are assigned to other agents based on their specialized capabilities (e.g., using models with strong code generation abilities for implementation steps while employing models with superior planning capabilities for architectural design). This specialization maximizes the quality of outputs while minimizing the likelihood of failures.

The coordination system also incorporates evaluation mechanisms that assess the quality of agent outputs, enabling the system to request refinements when necessary. This phase creates a robust cognitive framework that enables sophisticated reasoning and resilient multi-model collaboration, thereby forming the computational foundation for generating complex pipelines.

\subsection{Phase 3: Resilient Execution and Artifact Management} \label{subsec:Resilient Execution}

The final phase executes the plan while ensuring resilience and managing artifacts. This phase is implemented across~\cref{alg:AutoStreamPipe} (lines 20-35) and~\cref{alg:ESR}, with the resilient model handler layer in Figure~\ref{fig:AutoStreamPipe-details} directly corresponding to~\cref{alg:ESR}. 

\textbf{Step 1: Resilient model handler.} The resilient model handler employs four key strategies. Multi-agent model rotation is implemented through the \texttt{SwitchToNextModel} function (line~\ref{alg:es7}). The jitter-induced exponential backoff is defined in lines~\ref{alg:es5}-\ref{alg:es6}, using the formula shown in line~\ref{alg:es5-1}. The handling of the rate limit and graceful degradation are managed through the handling of conditional errors (lines~\ref{alg:es5}-\ref{alg:es8}) and the generation of the fallback result (line~\ref{alg:es9}).

\textbf{Step 2: Step-by-Step Plan Execution. }
The step-by-step plan execution ensures that steps are performed only when their dependencies are satisfied (Algorithm~\ref{alg:AutoStreamPipe}, lines \ref{alg:ASP16}-\ref{alg:ASP18}), allowing each phase to build upon the outputs of its prerequisites. Each step, such as \texttt{analyze\_complexity} or \texttt{design}, incorporates retry mechanisms to handle transient failures (line~\ref{alg:ASP19}). The results of each step are saved methodically (line~\ref{alg:ASP20}), creating a persistent record of the process. When results include Java code, as is common in steps like \texttt{generate\_pipeline}, the system extracts the code for separate storage to facilitate deployment (lines~\ref{alg:ASP21}-\ref{alg:ASP23}). Each step is marked as completed upon successful execution (line~\ref{alg:ASP23}), updating the execution plan's state to reflect progress. The visual execution process, shown in Figure~\ref{fig:AutoStreamPipe-details}, illustrates the progression, with arrows indicating dependencies between steps. This structured execution ensures that complex pipeline generation proceeds logically, with each step receiving the necessary inputs from prior operations.

\textbf{Step 3: Artifact Management.} We organize generation outputs into coherent, usable deliverables.
The SP pipeline is extracted from relevant steps, particularly \texttt{generate\_pipeline}, and formatted according to language-specific conventions. The step results are saved in a structured JSON format, providing a complete record for quality evaluation or debugging. The final response is synthesized (Algorithm~\ref{alg:AutoStreamPipe},~\ref{alg:ASP24}) into a narrative detailing the pipeline architecture, implementation, and operational aspects. A session summary is created (line~\ref{alg:ASP25}), cataloging all generated artifacts with explanations of their purposes and relationships. Additionally, the memory of the interaction is saved for future reference (line~\ref{alg:ASP26}), enabling the system to build upon prior experiences when handling similar requests. This process transforms the raw outputs of individual steps into a cohesive collection of interrelated resources that collectively satisfy the user's request.

Finally, the output generation produces deliverables tailored to the user's needs. These include an SP pipeline solution, consisting of executable code, configuration files, supporting resources, and comprehensive documentation that covers architecture, implementation, deployment, and operations. All artifacts are returned as the final output (Algorithm~\ref{alg:AutoStreamPipe}, line~\ref{alg:ASP27}), marking the transformation of a natural language request into a production-ready implementation through multi-agent collaboration and resilient execution.

\subsection{Integration and System Flow}

The complete \textsc{AutoStreamPipe} system integrates all three phases into a seamless workflow, transforming natural language queries into deployable SP pipelines. Algorithm~\ref{alg:AutoStreamPipe} serves as the central controller, orchestrating the entire process from initialization to final output generation. It begins with establishing the model infrastructure, including backup models for resilience, then proceeds through query analysis, graph building, plan execution, and artifact management.

\begin{figure*}[h]
    \centering
    % Row 1: Two equal-width subfigures
    \begin{subfigure}[b]{0.49\linewidth} % Slightly reduced to avoid overflow
        \centering
        \includegraphics[width=\linewidth]{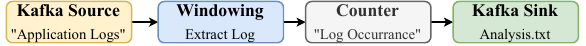}
        \caption{Log aggregator application (simple)}
        % \vspace{0.6em}
        \label{fig:image1}
    \end{subfigure}
    \hfill
    \begin{subfigure}[b]{0.49\linewidth} % Slightly reduced to avoid overflow
        \centering
        \includegraphics[width=\linewidth]{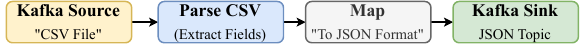}
        \caption{CSV data transformation (simple)}
        % \vspace{0.6em}
        \label{fig:image2}
    \end{subfigure}
    
    % \vspace{0.8em}
    % Row 2: Left column with one image; right column divided into two rows for two images
    \begin{subfigure}[b]{0.49\linewidth} % First column with one image
        \centering
        \includegraphics[width=\linewidth]{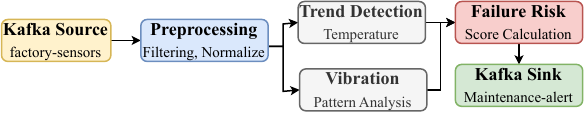}
        \caption{Industrial equipment predictive maintenance (medium)}
        \vspace{0.6em}
        \label{fig:image3}
    \end{subfigure}
    \hfill
    \begin{minipage}[b]{0.49\linewidth} % Second column split into two rows
        \begin{subfigure}[b]{\linewidth} % Top image in the second column
            \centering
            \includegraphics[width=\linewidth]{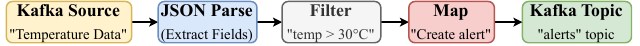}
            \caption{Temperature monitoring \cite{huan2021theoretical} (medium)}
            \vspace{0.4em}
            \label{fig:image4}
        \end{subfigure}
        \begin{subfigure}[b]{\linewidth} % Bottom image in the second column
            \centering
            \vspace{0.3em}
            \includegraphics[width=\linewidth]{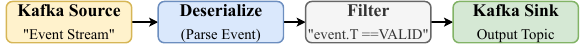}
            
            \caption{Event filtering (medium)}
            \vspace{0.4em}
            \label{fig:image5}
        \end{subfigure}

    \end{minipage}
    
    \begin{subfigure}[b]{0.63\linewidth} % Adjusted slightly to avoid overflow
        \centering
        \includegraphics[width=\linewidth]{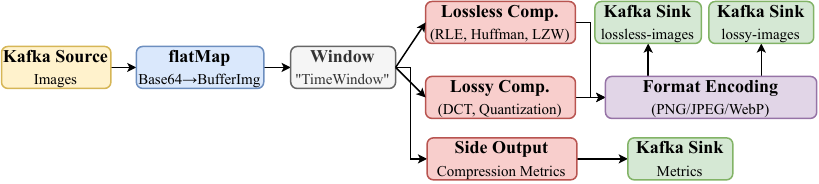}
        \caption{Image compression \cite{pandimurugan2022hybrid} (complex)}
        % \vspace{0.6em}
        \label{fig:image5}
    \end{subfigure}
    \hfill
    \begin{subfigure}[b]{0.35\linewidth}
        \centering
        \includegraphics[width=\linewidth]{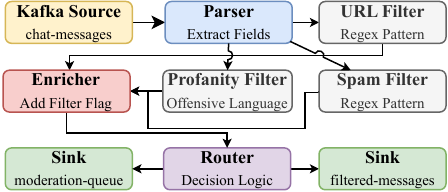}
        \caption{Chat message filtering \cite{1532048} (complex)}
        % \vspace{0.6em}
        \label{fig:image6}
    \end{subfigure}
    
    \caption{Overview of various data processing pipelines used in different applications, showcasing their architecture}
    \label{fig:main_figure}
\end{figure*}

The \textsc{AutoStreamPipe} architecture significantly advances automated SP pipeline generation, bridging the gap between natural language expressions of intent and production-ready implementation. By combining structured reasoning through hypergraphs, multi-agent collaboration across model providers, and resilience mechanisms, the system achieves a level of reliability and capability that exceeds traditional single-model approaches. The detailed algorithms formally define the system's operation, demonstrating how each visual component in Figure~\ref{fig:AutoStreamPipe-details} is implemented in practice.

%%%%%%%%%%%%%%%%%%%%%%%%%%%%%%%%%%%%%%%%%%%%%%%%%%%%%%%%%%%%%%
\section{Performance Evaluation} \label{sec:performance}

\subsection{Experimental Setup}
\label{subsec:exp_setup}

We comprehensively evaluated the \textsc{AutoStreamPipe}, testing its planning and resilient execution capabilities. All experiments were conducted on a system equipped with an Intel Core i7-13700K processor (16 cores, 3.4 GHz) and 64GB of RAM, running Ubuntu 22.04 LTS. \textsc{AutoStreamPipe} was developed in Python (v3.12.7) and integrates LangChain (v0.3.11) for LLM support. For testing the generated pipelines, we used the latest stable versions of Apache Flink (v1.20.1), Storm (v2.8.0), and Spark (v3.5.5). We evaluated two types of LLM: Codestral Mamba \cite{mistral_models_overview} and llama-3.3 \cite{groq_models} as our Open Source (OS) option, and ChatGPT-4o-mini \cite{openai_pricing} and Claude-Haiku \cite{anthropic_pricing} as our Closed Source (CS) option, with detailed comparisons available in our evaluation table. The \textsc{AutoStreamPipe} codebase comprises about 4,000 lines of Python for the framework implementation and 15,000 lines of Java for the applications.

\subsubsection{Dataset and Queries.}
In our paper, we develop a benchmark suite that contains eight diverse SP applications specifically designed for pipeline design evaluation. This suite includes self-defined applications + word count \cite{sun2022energy,li2022dynamic,9644756} (see Figure~\ref{fig:main_figure}). For each application, we define two query types: full- and partial-information queries. For example, the word count application is illustrated by the full information query in the box \ref{box:wordcount}, where all relevant information is provided in the prompt, and the partial information in the box ~\ref{box:red}, where some details are missing. All queries are available in our repository\footnote{\label{fn:github} \url{https://github.com/Anonymous0-0paper/SWG/blob/master/Query_docs.txt}}.
All benchmarks are available on the \textsc{AutoStreamPipe} GitHub repository\footnote{\url{https://github.com/Anonymous0-0paper/SWG}}. We carefully selected these applications to design their pipelines, providing a balanced mix of three simple, three medium, and two complex pipelines that target commonly used SPEs, along with a custom data generator for each pipeline. To account for variability in the generation process, we produced each pipeline five times for each of the three SPEs (Flink, Storm, and Spark). %This approach resulted in 120 generated pipelines, which we then rigorously evaluated against the ground truth to determine the precision of \textsc{AutoStreamPipe}.
% \begin{elegantbox}[label=lst:red]%
%   {Word Count Pipeline (Partial Description Version)}
% Create a Flink streaming application for text processing:
% \begin{itemize}[wide]
%   \item \textbf{Source:} Write to Kafka topic
%   \item \textbf{Input format:} Plain text messages
%   \item \textbf{Process:} Split messages into words by whitespace and count frequencies
%   \item \textbf{Windowing:} Aggregate every 30 seconds
%   \item \textbf{Output:} Local file at \texttt{word-counts.txt} 
% \end{itemize}
% \end{elegantbox}
\begin{infobox}\footnotesize
\caption{Word Count Pipeline (Partial Description Version)}
\label{box:red}
\begin{redbox}[frametitle={Word Count Pipeline (Partial Description Version)}]
Create a Flink streaming application for text processing:
\begin{itemize}[wide]
  \item \textbf{Source:} Write to Kafka topic
  \item \textbf{Input format:} Plain text messages
  \item \textbf{Process:} Split messages into words by whitespace and count frequencies
  \item \textbf{Windowing:} Aggregate every 30 seconds
  \item \textbf{Output:} Local file at \texttt{word-counts.txt} 
\end{itemize}
\end{redbox}
\end{infobox}

\begin{table}[b]
\centering
\setlength{\extrarowheight}{0pt}
\setlength{\aboverulesep}{0pt}
\setlength{\belowrulesep}{0pt}
\caption{LLM Models. MTok: Millions of Tokens, (I/O): Input / Output Token}
\label{tab:evaluationModels}
\resizebox{\linewidth}{!}{%
\LARGE
\begin{tabular}{ccccc} 
\toprule
\rowcolor[rgb]{0.812,0.812,0.812} \textbf{Type} & \textbf{Provider} & \textbf{LLM Model} & \textbf{API} & \textbf{Price (\textdollar/(I/O)MTok)} \\ 
\midrule
\multirow{2}{*}{OS} & Mistral AI & Codestral Mamba & \begin{tabular}[c]{@{}c@{}}open-codestral\\-mamba\end{tabular} & - \\ 
\cdashline{2-5}
 & Meta & llama-3.3 & \begin{tabular}[c]{@{}c@{}}llama-3.3-70b\\-versatile\end{tabular} & - \\ 
\hline\hline
\multirow{2}{*}{CS} & Anthropic & Claude & \begin{tabular}[c]{@{}c@{}}\textcolor[rgb]{0.067,0.094,0.153}{claude-3-5-haiku}\\\textcolor[rgb]{0.067,0.094,0.153}{-20241022}\end{tabular} & \$0.80 / \$4.00 \\ 
\cdashline{2-5}
 & Open AI & gpt-4o-mini & \begin{tabular}[c]{@{}c@{}}gpt-4o-mini\\-2024-07-18\end{tabular} & \$0.15 / \$0.60 \\
\bottomrule
\end{tabular}
}
\end{table}
\subsubsection{Baseline Systems.}
We compared our system against the following alternatives:
\begin{itemize}[leftmargin=*]
   \item \textbf{Base-LLM}: Direct queries to the LLM without HGoT, planning, and resilience features.
   \item \textbf{CoT Planning}: Our system with query analysis and CoT planning without resilience mechanisms.
   \item \textbf{GoT based}: Standard GoT implementation without hypergraph capabilities and our query analyzer.
   \item \textbf{\textsc{AutoStreamPipe} (ASP)}: Complete implementation with HGoT, query analysis, planning, and resilient execution.
\end{itemize}

\subsection{Evaluation Metrics}
\label{subsec:metrics}

We evaluated our system using several key metrics. In some cases, we relied on unit tests because the advanced capabilities of LLM often exceed the scope of standard automated evaluation metrics for general natural language tasks \cite{chang2024survey,guo2023evaluating}.

\textbf{Processing Time}. Total time required to process a query, measured in seconds.
\textbf{Response Completeness}. Percentage of query requirements addressed in the response.

\begin{table}[b]
\centering
% \scriptsize
\setlength{\extrarowheight}{0pt}
\setlength{\aboverulesep}{0pt}
\setlength{\belowrulesep}{0pt}
\caption{Error-free Score (EFS) comparison. PL: Pipeline}
\label{tab:efs_comparison}
\resizebox{\linewidth}{!}{%
\begin{tabular}{cccccc} 
\toprule
\rowcolor[rgb]{0.812,0.812,0.812} \textbf{PL Type} & \textbf{SPE} & \textbf{Base-LLM} & \textbf{CoT Planning} & \textbf{GoT Based} & \textbf{AutoStreamPipe} \\ 
\midrule
\multirow{3}{*}{Simple} & Flink & 0.36 & 0.41 & 0.51 & \textbf{1} \\
 & Storm & 0.54 & 0.56 & 0.78 & \textbf{1} \\
 & Spark & 0.47 & 0.51 & 0.65 & \textbf{0.94} \\ 
\hdashline
\rowcolor[rgb]{0.867,0.961,0.906} \multicolumn{2}{c}{\textbf{\textbf{\textbf{\textbf{Average}}}}} & 0.46 & 0.49 & 0.65 & \textbf{0.98} \\ 
\hline\hline
\multirow{3}{*}{Medium} & Flink & 0.26 & 0.32 & 0.4 & \textbf{0.68} \\
 & Storm & 0.27 & 0.28 & 0.5 & \textbf{0.73} \\
 & Spark & 0.32 & 0.36 & 0.52 & \textbf{0.69} \\ 
\hdashline
\rowcolor[rgb]{0.867,0.961,0.906} \multicolumn{2}{c}{\textbf{\textbf{Average}}} & 0.28 & 0.32 & 0.47 &\textbf{ 0.7} \\ 
\hline\hline
\multirow{3}{*}{Complex} & Flink & 0.18 & 0.24 & 0.36 & \textbf{0.63}\\
 & Storm & 0.27 & 0.27 & 0.44 & \textbf{0.6} \\
 & Spark & 0.24 & 0.26 & 0.39 & \textbf{0.54} \\ 
\hdashline
\rowcolor[rgb]{0.867,0.961,0.906} \multicolumn{2}{c}{\textbf{Average}} & 0.23 & 0.26 & 0.4 & \textbf{0.59} \\
\bottomrule
\end{tabular}
}
\end{table}

\textbf{Error-free Score (EFS).} To comprehensively assess the quality of a generated pipeline, we introduce a metric called the Error-Free Score (EFS). This metric evaluates the pipeline's accuracy (meaning how often the model generates correct pipelines) and quality by considering multiple errors that can occur during its creation and execution. The EFS is designed to provide a more nuanced and fine-grained assessment of pipeline quality, surpassing simple measures such as compilation success rates. The EFS is calculated using the following equation:
\begin{equation}
    \text{EFS} = \frac{1}{3} \left( \frac{1}{1+S} + \frac{1}{1+L} + \frac{1}{1+R} \right)
\end{equation}
where $S$ is the number of syntax errors in the generated pipeline, $L$ means the number of logical errors (correct syntax but incorrect algorithm/logic), and $R$ is the number of runtime errors when executing the code (e.g., null pointer dereferencing).
The EFS produces a score between 0 and 1, where 1 represents an entirely error-free pipeline. This metric provides a fine-grained assessment of pipeline quality beyond simple compilation success rates.

%\textbf{Error-free Score.} In 
\subsection{Evaluation Results} \label{subsec:results}
\begin{figure*}[!t]
    \centering
    %%% First Row %%%
    \begin{subfigure}[t]{0.32\textwidth}
        \centering
        \includegraphics[width=\linewidth]{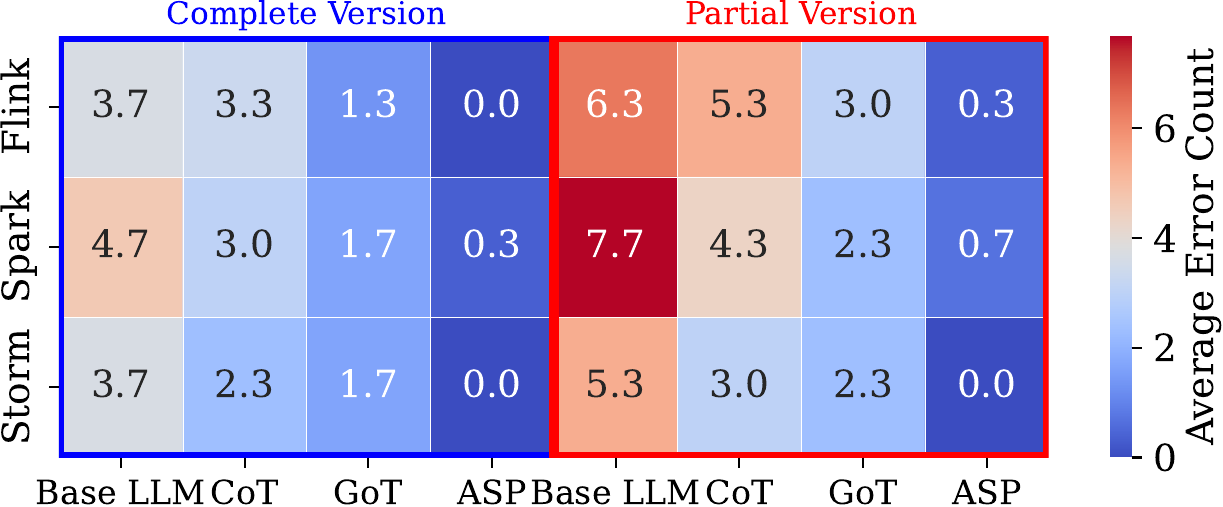}
        \vspace{-0.5cm}
        \caption{\scriptsize Syntax errors on simple pipelines}
        \label{fig:exp1}
    \end{subfigure}
    \hfill
    \begin{subfigure}[t]{0.32\textwidth}
        \centering
        \includegraphics[width=\linewidth]{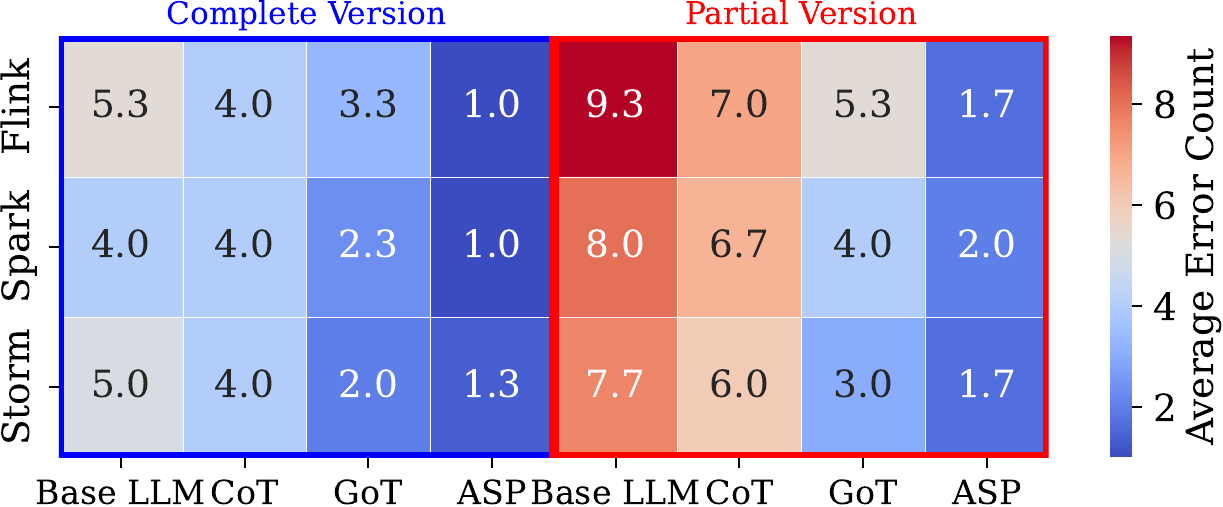}
        \vspace{-0.5cm}
        \caption{\scriptsize Syntax errors on medium pipelines}
        \label{fig:exp2}
    \end{subfigure}
    \hfill
    \begin{subfigure}[t]{0.32\textwidth}
        \centering
        \includegraphics[width=\linewidth]{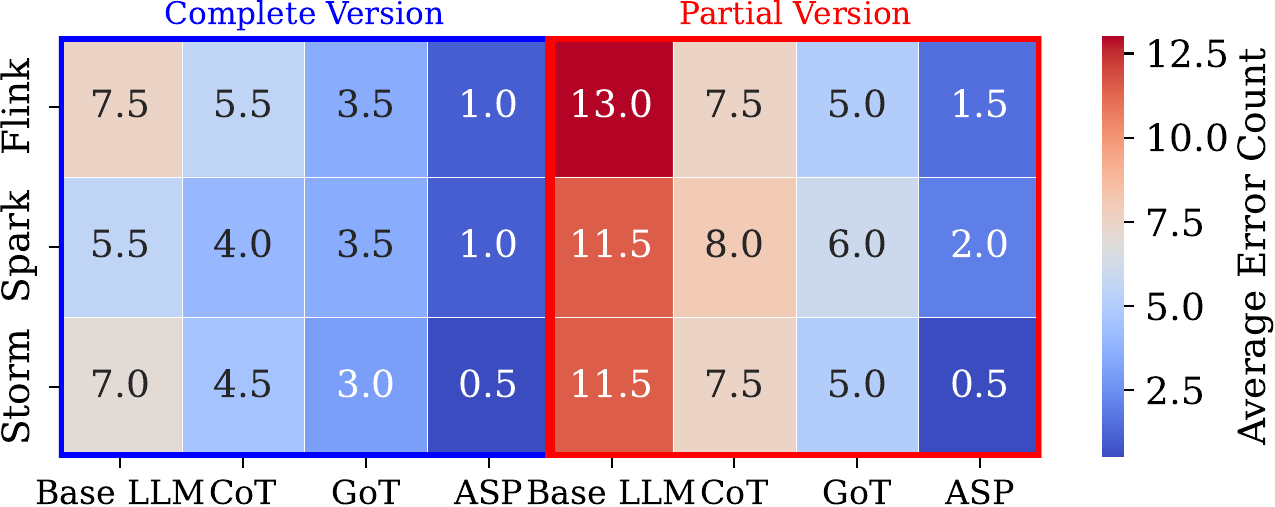}
        \vspace{-0.5cm}
        \caption{\scriptsize Syntax errors on complex pipelines}
        \label{fig:exp3}
    \end{subfigure}
    
    % \vspace{0.4cm}
    %%% Second Row %%%
    \begin{subfigure}[t]{0.32\textwidth}
        \centering
        \includegraphics[width=\linewidth]{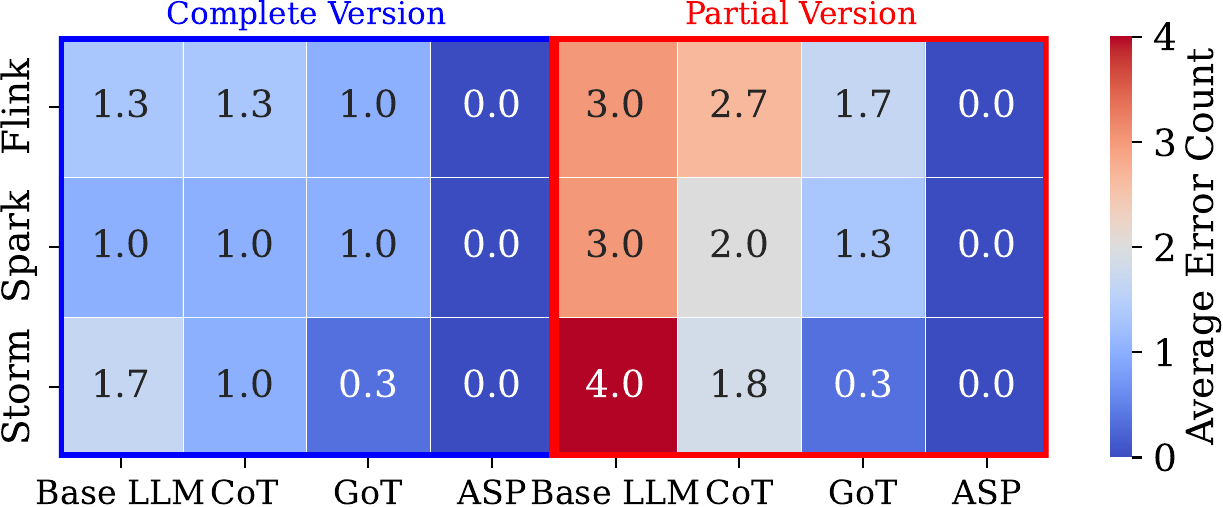}
        \vspace{-0.5cm}
        \caption{\scriptsize Logic errors on simple pipelines}
        \label{fig:exp4}
    \end{subfigure}
    \hfill
    \begin{subfigure}[t]{0.32\textwidth}
        \centering
        % Replace with your fifth experiment image
        \includegraphics[width=\linewidth]{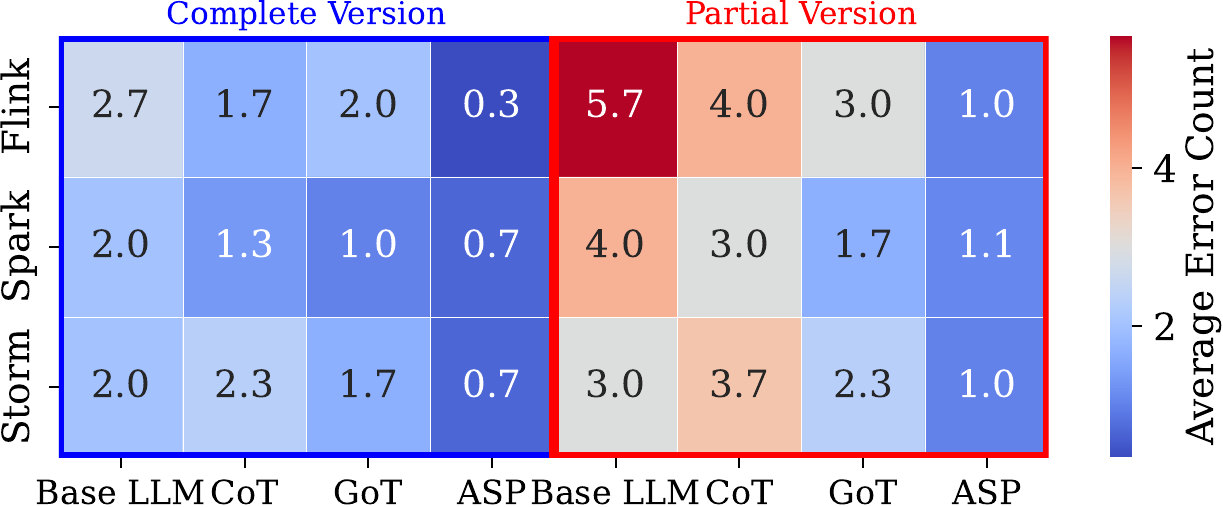}
        \vspace{-0.5cm}
        \caption{\scriptsize Logic errors on medium pipelines}
        \label{fig:exp5}
    \end{subfigure}
    \hfill
    \begin{subfigure}[t]{0.32\textwidth}
        \centering
        % Replace with your sixth experiment image
        \includegraphics[width=\linewidth]{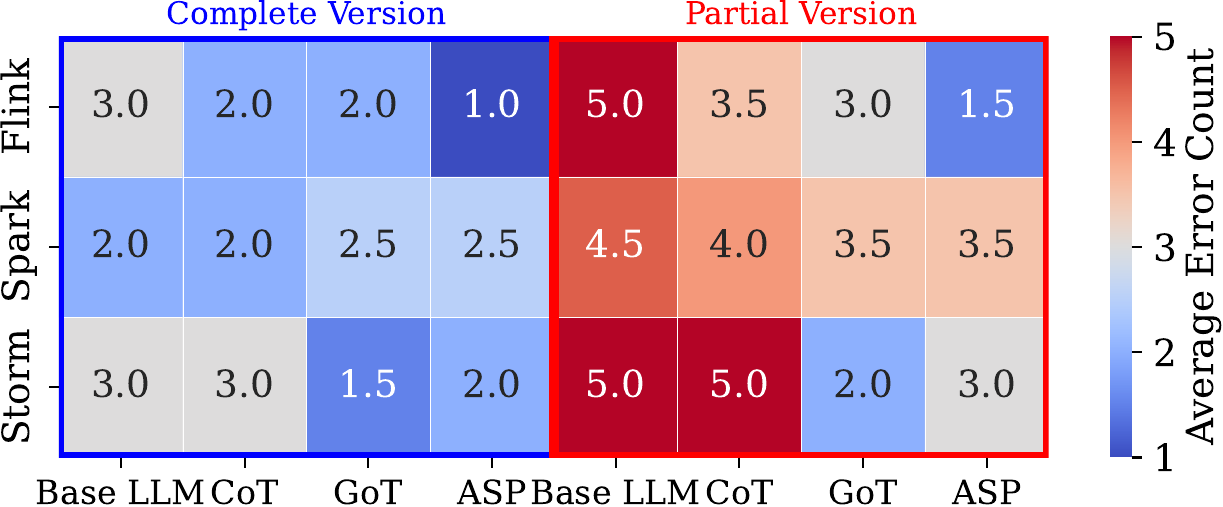}
        \vspace{-0.5cm}
        \caption{\scriptsize Logic errors on complex pipelines}
        \label{fig:exp6}
    \end{subfigure}
    
    % \vspace{0.4cm}
    %%% Third Row %%%
    \begin{subfigure}[t]{0.32\textwidth}
        \centering
        % Replace with your seventh experiment image
        \includegraphics[width=\linewidth]{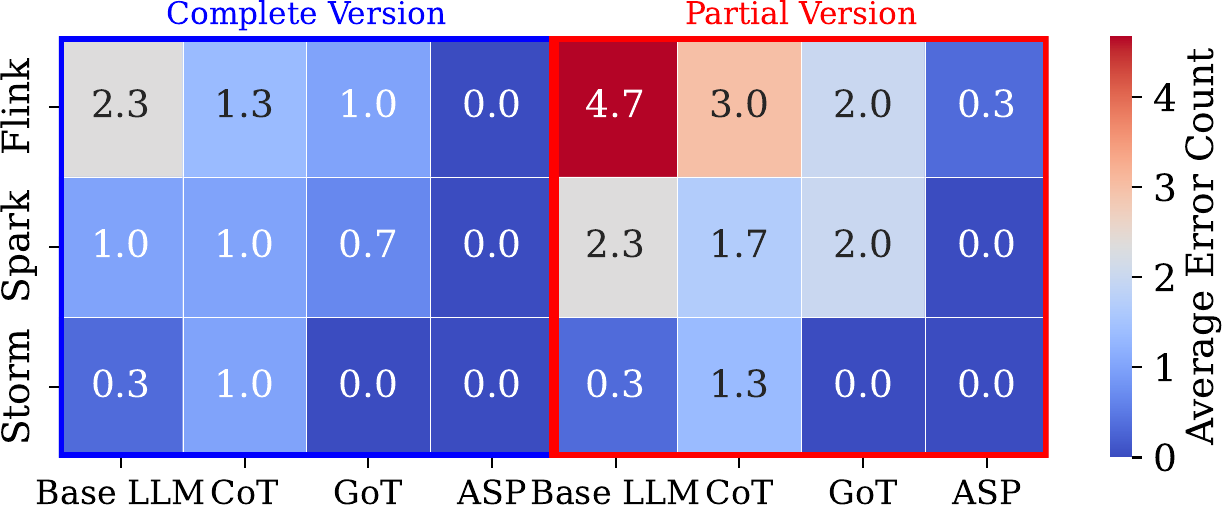}
        \vspace{-0.5cm}
        \caption{\scriptsize Runtime errors on simple pipelines}
        \label{fig:exp7}
    \end{subfigure}
    \hfill
    \begin{subfigure}[t]{0.32\textwidth}
        \centering
        % Replace with your eighth experiment image
        \includegraphics[width=\linewidth]{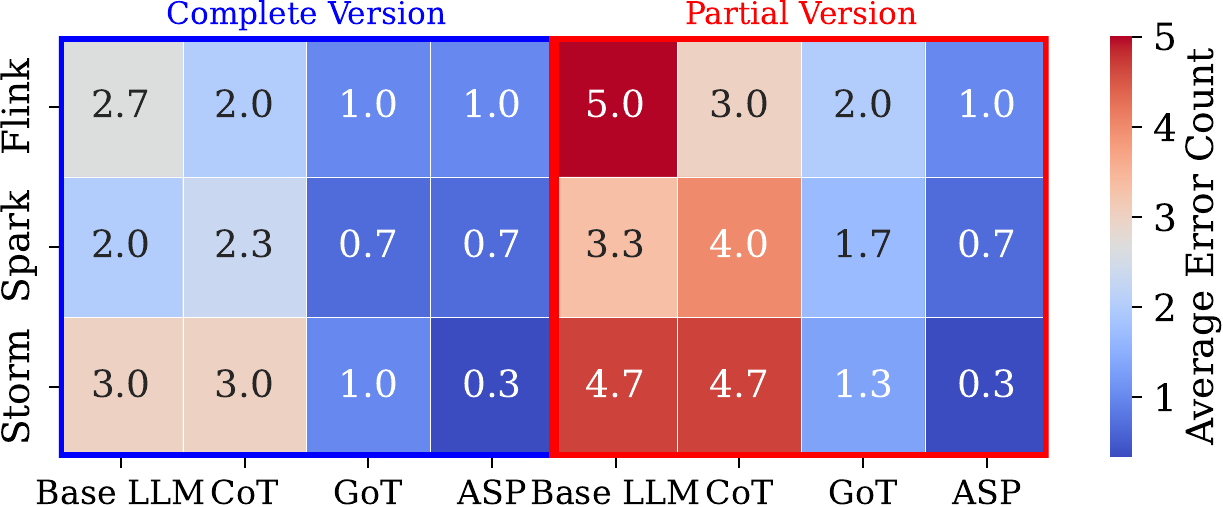}
        \vspace{-0.5cm}
        \caption{\scriptsize Runtime errors on medium pipelines}
        \label{fig:exp8}
    \end{subfigure}
    \hfill
    \begin{subfigure}[t]{0.32\textwidth}
        \centering
        % Replace with your ninth experiment image
        \includegraphics[width=\linewidth]{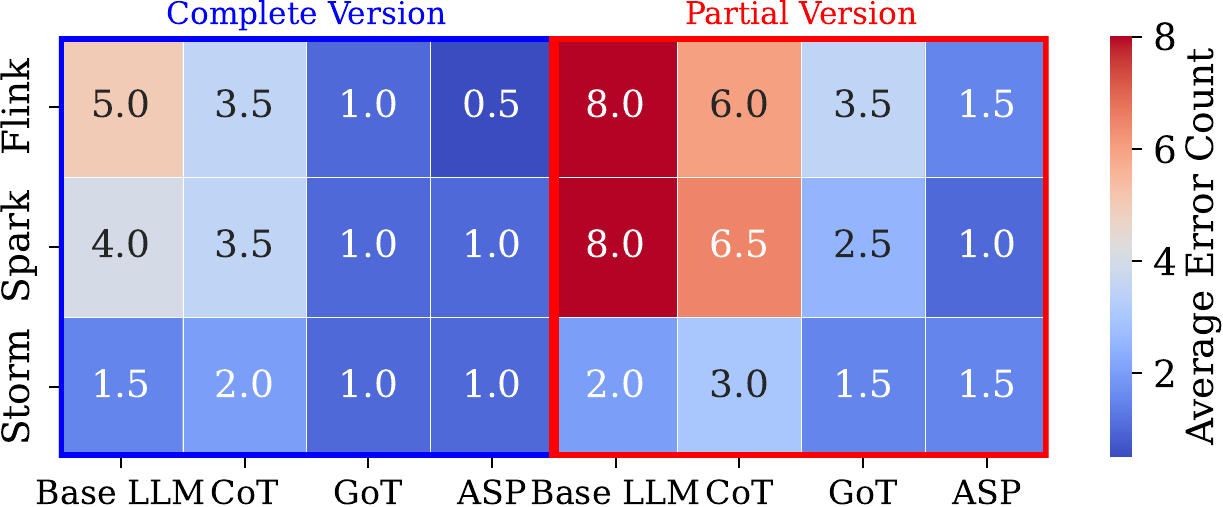}
        \vspace{-0.5cm}
        \caption{\scriptsize Runtime errors on complex pipelines}
        \label{fig:exp9}
    \end{subfigure}
    
    % \vspace{0.5cm}
    \caption{Comparison of average error counts (AEC) in syntax, logic, and error across different pipeline complexity levels (simple, medium, and complex) on complex and partial information versions for four distinct approaches (Base LLM, CoT, GoT, and \textsc{AutoStreamPipe} (ASP)). These plots illustrate how error frequencies vary with increasing pipeline complexity and approach, providing insights into performance differences.}
    \label{fig:comparison_SPE_average_Error}
\end{figure*}

\textbf{Error-Free Score}. Table \ref{tab:efs_comparison} compares EFS in three pipeline complexities (Simple, Medium, Complex) and three SPEs (Flink, Storm, Spark). Each row displays the EFS for four approaches: Base-LLM, CoT Planning, GoT-Based, and \textsc{AutoStreamPipe}, with an average value at the bottom of each complexity group.

\textsc{AutoStreamPipe} reaches an average EFS of 0.98 for simple pipelines, indicating near-perfect performance. This high score is due to the pipeline's simplicity, which minimizes the occurrence of syntax, logic, and runtime errors. In contrast, CoT Planning achieves an EFS of 0.65 and Base-LLM 0.46, highlighting the limitations of these methods in handling even simple tasks without error. The significant improvement in \textsc{AutoStreamPipe} is primarily due to the HGoT and advanced query analysis, which enhance the accuracy and consistency of the generated pipelines.
In medium pipelines, \textsc{AutoStreamPipe} maintains a high average EFS (0.73), significantly outperforming GoT Based (0.54), CoT Planning (0.44), and Base-LLM (0.36). This performance difference arises because medium-complexity pipelines introduce more intricate logical relationships and data transformations, which traditional methods struggle to capture accurately. The HGoT's multi-way reasoning and robust execution planning in \textsc{AutoStreamPipe} enable it to handle these challenges more effectively.
Performance declines for all approaches as pipeline complexity increases. For complex pipelines, \textsc{AutoStreamPipe} achieves an average EFS of 0.59. Although this score is lower than for simpler pipelines, it remains substantially higher than CoT Planning (0.42) and Base-LLM (0.23–0.30 range). The reduced performance is expected due to the inherent challenges of managing complex interdependencies. However, the structured reasoning of the HGoT and the resilient multi-agent execution in \textsc{AutoStreamPipe} ensure that it still achieves a 50\% improvement over CoT Planning and a 64\% improvement over Base-LLM.
Overall, \textsc{AutoStreamPipe} delivers the best results in every scenario due to HGOT and our query analyzer. In the best case (Simple pipelines), it achieves an average EFS of 0.98, and even in the worst case (Complex pipelines), it reaches 0.59, a 50\% improvement over CoT Planning and a 64\% improvement over the Base-LLM.

\begin{figure}
    \centering
    \includegraphics[width=1\linewidth]{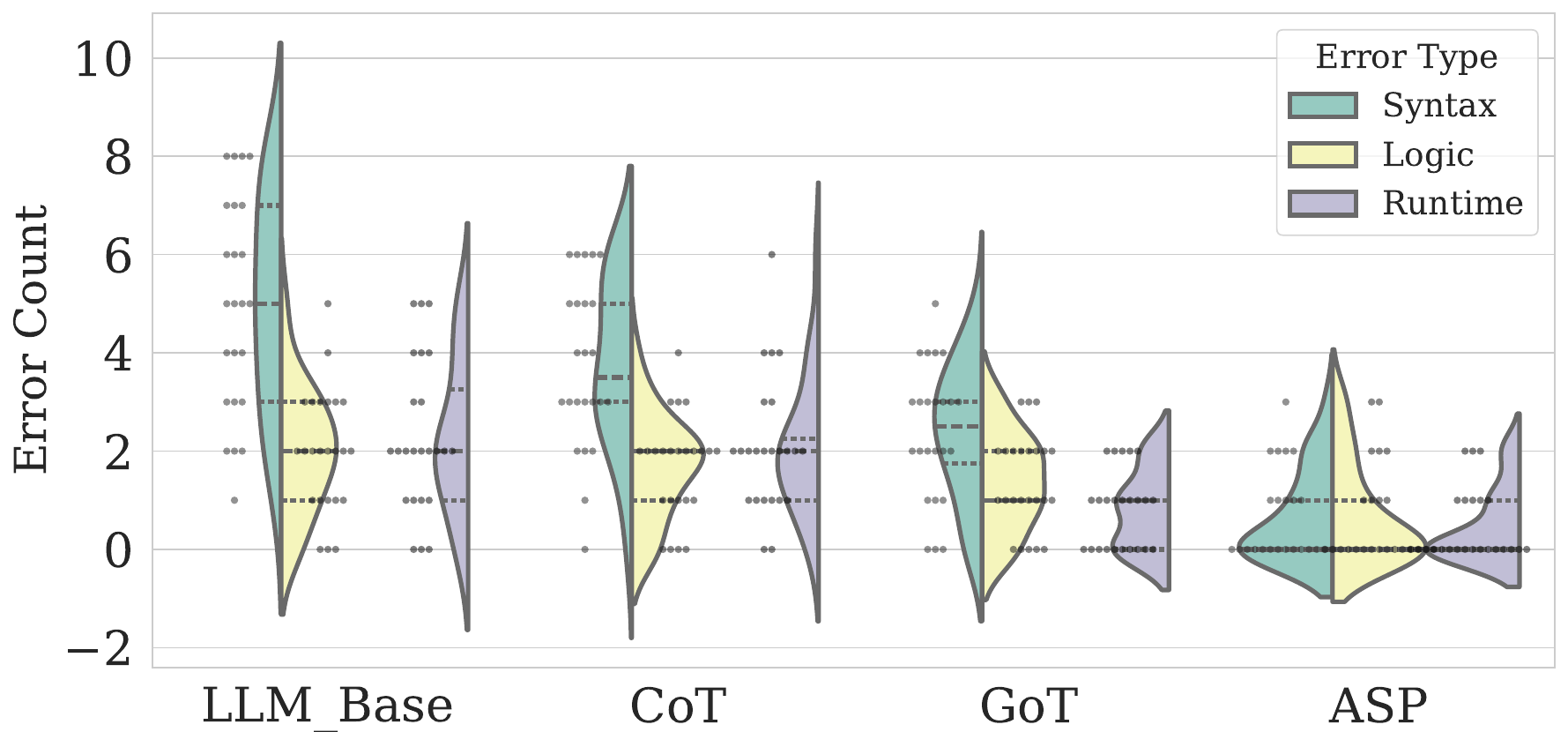}
    \caption{Distribution of error types across approaches}
    \label{fig: Distribution of Error Types Across Approaches}
\end{figure}

\textbf{Error Distribution and Composition.} Figures~\ref{fig:comparison_SPE_average_Error} and \ref{fig: Distribution of Error Types Across Approaches} present a detailed analysis of error patterns across various pipeline generation approaches and SPEs, to show how well each method handles different types of errors. Figure~\ref{fig:comparison_SPE_average_Error} visualizes the average counts of syntax, logic, and runtime errors across three levels of pipeline complexity (simple, medium, complex) and two input settings: complete and partial information. Across all configurations, we observe a consistent trend: as we move from the Base LLM to the proposed \textsc{AutoStreamPipe (ASP)}, the average error counts decrease substantially. This improvement is especially evident under the partial information setting, where approaches must infer or complete missing pipeline specifications. Syntax errors are the most dominant across all models, particularly in Base LLM and CoT, which lack structured planning mechanisms. These errors are mitigated more effectively by ASP due to its robust multi-agent coordination and planning capabilities. In contrast, logic errors show a moderate reduction as we move from simpler to more advanced methods. ASP’s enhanced context understanding and error-aware planning significantly reduce these issues, particularly in medium- to complex-sized pipelines.

Runtime errors, while fewer in number, represent critical failures in executable pipelines. These errors are persistently challenging for all methods, but ASP demonstrates a clear advantage, particularly in complex pipelines with partial information, where it maintains the lowest runtime error counts.
Importantly, the gap between the complete and partial versions increases with the complexity of the pipeline. All baseline models degrade significantly under partial inputs, whereas ASP maintains low error rates and high stability, demonstrating its robustness in uncertain and ambiguous input scenarios.

Figures~\ref{fig:exp1}-\ref{fig:exp9} complement this analysis by presenting the average error composition for three SPEs. Here, Syntax errors dominate the total error count in Base-LLM because this method lacks structured reasoning and robust error handling. Logic and runtime errors are less frequent but significant, indicating that even basic pipeline generation methods can occasionally produce syntactically correct but logically flawed code. As the approaches become more advanced, the \textsc{AutoStreamPipe} method achieves the lowest overall error counts, with balanced reductions across all error types. The \textsc{AutoStreamPipe} minimizes syntax errors through improved planning and addresses logical inconsistencies through enhanced reasoning, as well as runtime issues via resilient execution strategies.
\begin{figure}[b]
    \centering
    \includegraphics[width=1\linewidth]{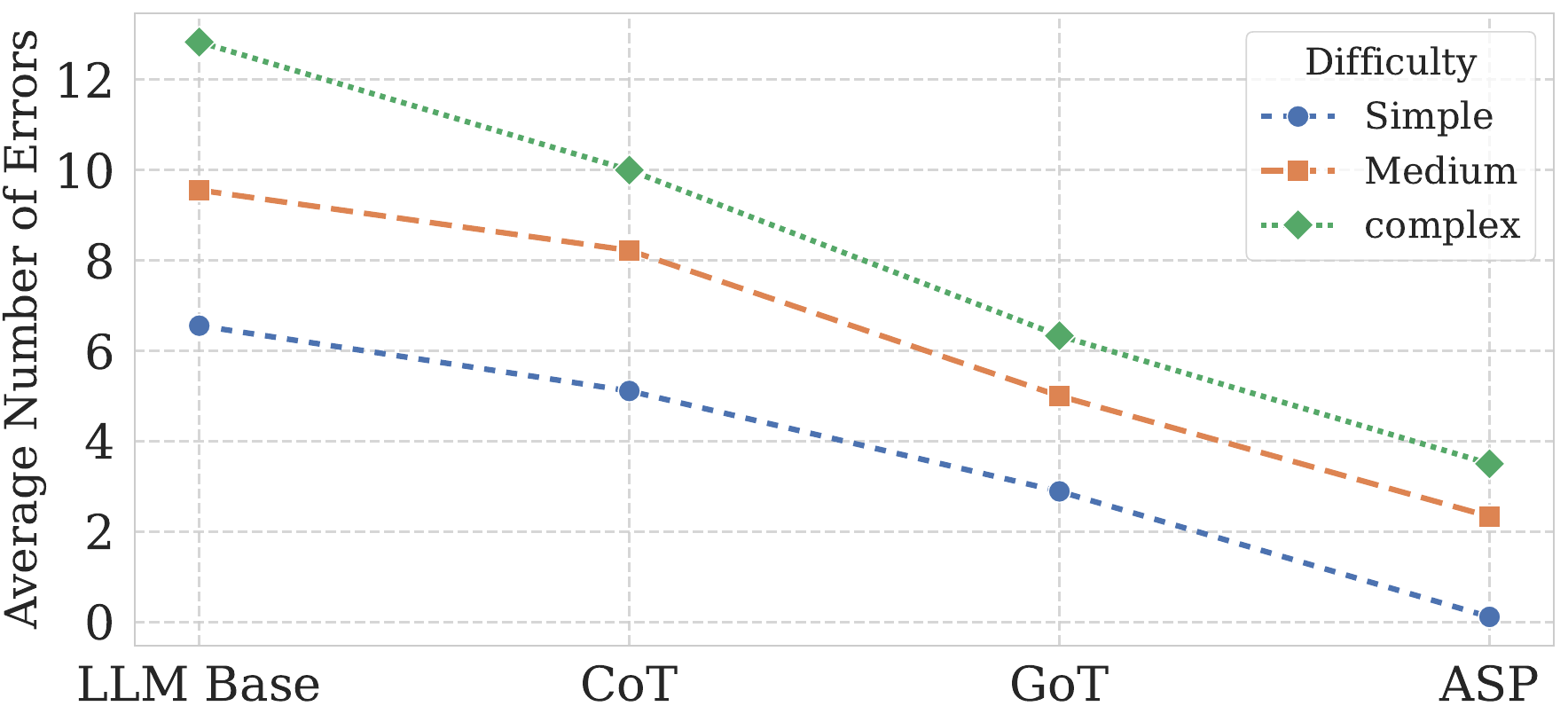}
    \caption{Error reduction by difficulty level across approaches}
    \label{fig:fancy_difficulty_progression}
\end{figure}
Figures~\ref{fig:exp1}-\ref{fig:exp9}, and \ref{fig: Distribution of Error Types Across Approaches} underscore the effectiveness of the \textsc{AutoStreamPipe} in minimizing errors and highlight the importance of selecting an appropriate SPE to optimize pipeline quality.

\textbf{Error Reduction by Difficulty Level.} Figure~\ref{fig:fancy_difficulty_progression} illustrates the reduction in average errors across three pipeline difficulty levels. The \textsc{AutoStreamPipe} consistently achieves the lowest error rates across all difficulty levels, demonstrating its superior performance compared to the other methods.
This consistency is due to the \textsc{AutoStreamPipe}'s ability to adapt its reasoning and execution strategies according to pipeline complexity, unlike other methods that struggle as pipelines become more intricate.
As pipeline complexity increases, the average number of errors rises for all approaches, but the gap between Base LLM and \textsc{AutoStreamPipe} widens significantly, particularly for complex pipelines. This pattern highlights that basic methods are prone to generating errors when faced with complex data transformations and dependencies. \textsc{AutoStreamPipe}'s multi-agent coordination and hypergraph reasoning help maintain lower error rates. 
Intermediate approaches, such as CoT and GoT, demonstrate gradual improvements over the baseline, highlighting their incremental effectiveness. Furthermore, the consistent reduction in errors from Base LLM to \textsc{AutoStreamPipe} underscores the cumulative benefits of integrating advanced strategies into pipeline generation. 
Overall, Figure~\ref{fig:fancy_difficulty_progression} highlights the critical role of pipeline complexity in influencing error rates and demonstrates the robustness of \textsc{AutoStreamPipe} in achieving reliable performance across varying levels of complexity. These insights are crucial for understanding the practical applications of pipeline generation methods and selecting the most suitable strategy for various complexities.

\begin{figure}[t]
    \centering
    
    %%% First Row %%%
    \begin{subfigure}[t]{0.48\columnwidth} % Adjusted to fit column width
        \centering
        \includegraphics[width=\linewidth]{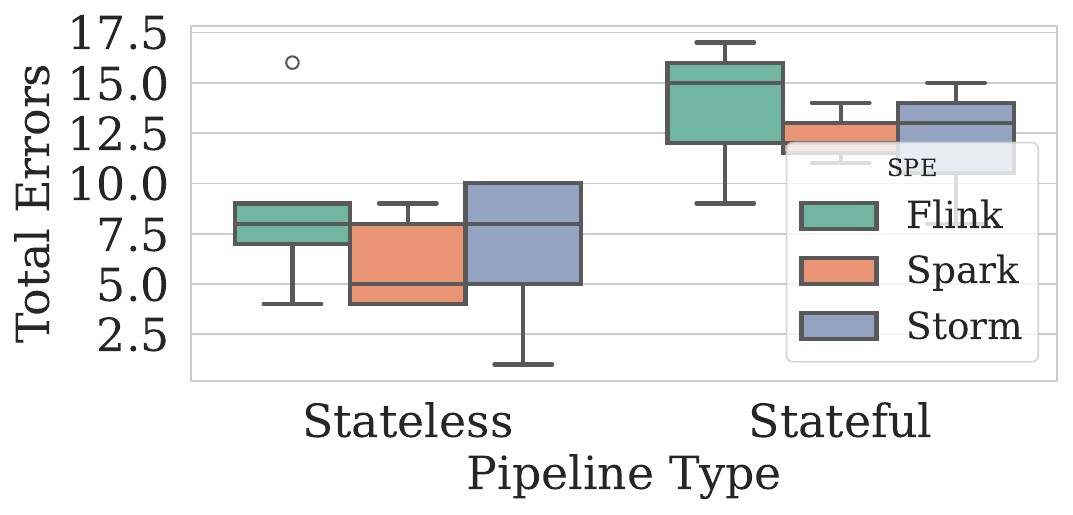}
        \vspace{-0.7cm}
        \caption{\scriptsize Base LLM pipeline quality}
        \label{fig:llm_base}
    \end{subfigure}
    \hfill
    \begin{subfigure}[t]{0.48\columnwidth} % Adjusted to fit column width
        \centering
        \includegraphics[width=\linewidth]{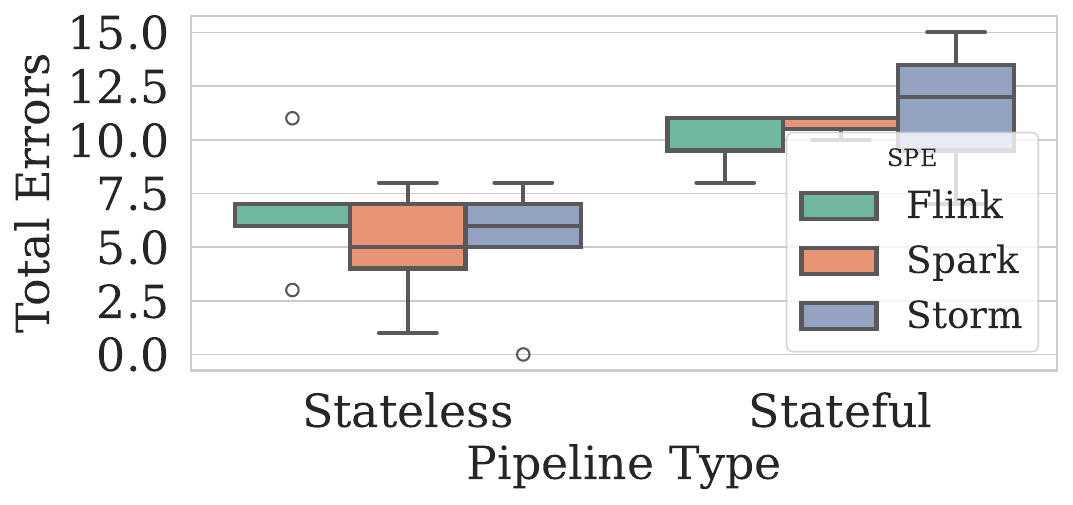}
        \vspace{-0.7cm}
        \caption{\scriptsize CoT pipeline quality}
        \label{fig:cot}
    \end{subfigure}

    %%% Second Row %%%
    \begin{subfigure}[t]{0.48\columnwidth} % Adjusted to fit column width
        \centering
        \includegraphics[width=\linewidth]{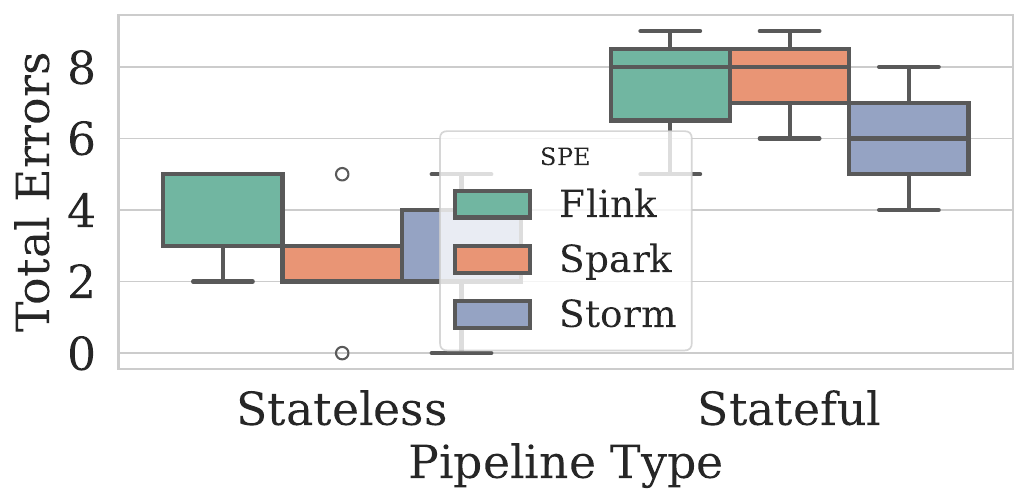}
        \vspace{-0.7cm}
        \caption{\scriptsize GoT pipeline quality}
        \label{fig:got}
    \end{subfigure}
    \hfill
    \begin{subfigure}[t]{0.48\columnwidth} % Adjusted to fit column width
        \centering
        \includegraphics[width=\linewidth]{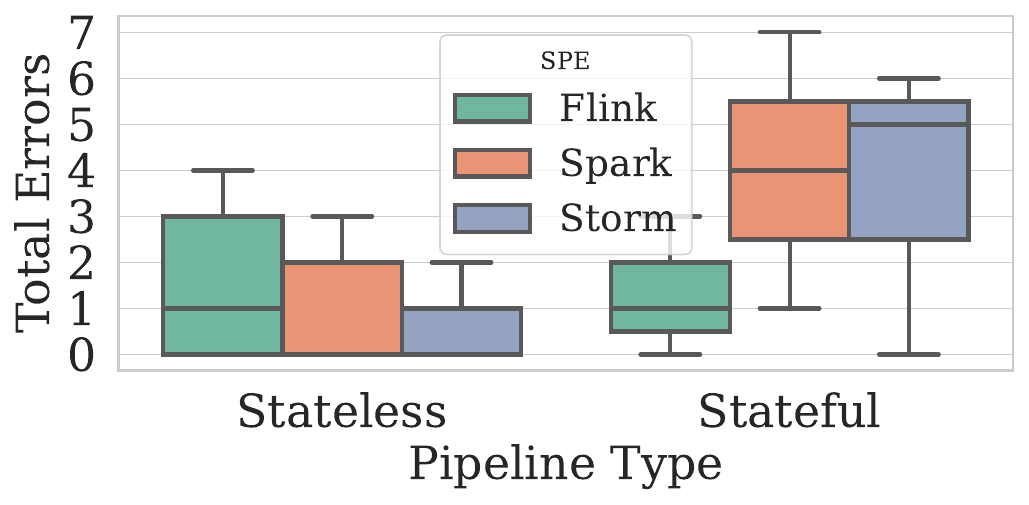}
        \vspace{-0.7cm}
        \caption{\scriptsize \textsc{AutoStreamPipe} pipeline quality}
        \label{fig:full}
    \end{subfigure}
    
    \vspace{0.5cm}  % Consistent spacing before the main caption
    \caption{Comparison of quality of pipelines across four distinct approaches for stateless and stateful pipelines.}
    \label{fig:comparison_errors_approaches}
\end{figure}

\textbf{Stateless and Stateful Pipeline Quality.}
The pipeline quality generated using \textsc{AutoStreamPipe} was evaluated using automated compilation tests. Lower total errors mean higher quality.
Figure~\ref{fig:comparison_errors_approaches} analyzes the quality of pipelines generated by four approaches: Base LLM, CoT, GoT, and \textsc{AutoStreamPipe} across two pipeline types (Stateless and Stateful) and three SPEs. The figure is divided into four subplots~(\ref{fig:llm_base}) -~(\ref{fig:full}), each representing the error distributions for a specific approach. Across all approaches, stateless pipelines consistently exhibit lower error rates than stateful pipelines, reflecting the added complexity of managing stateful operations. The Base LLM approach in Figure~(\ref{fig:llm_base}) exhibits the highest error rates, with significant variability and pronounced outliers, particularly in stateful pipelines. The CoT approach in Figure~(\ref{fig:cot}) demonstrates noticeable improvements, reducing median errors and variability. Meanwhile, GoT in Figure~(\ref{fig:got}) further refines performance, especially for stateful pipelines. The \textsc{AutoStreamPipe} in Figure~(\ref{fig:full}) achieves the best results, with the lowest median error counts, a tight interquartile range, and minimal outliers, indicating consistent and reliable performance. 
These findings demonstrate the effectiveness of \textsc{AutoStreamPipe} in mitigating errors, particularly in complex stateful scenarios, and highlight the importance of selecting an appropriate SPE to optimize pipeline quality.

\begin{figure}[t]
    \centering

    % First Subfigure
    \begin{subfigure}[t]{0.48\columnwidth}
        \centering
        \includegraphics[width=\linewidth]{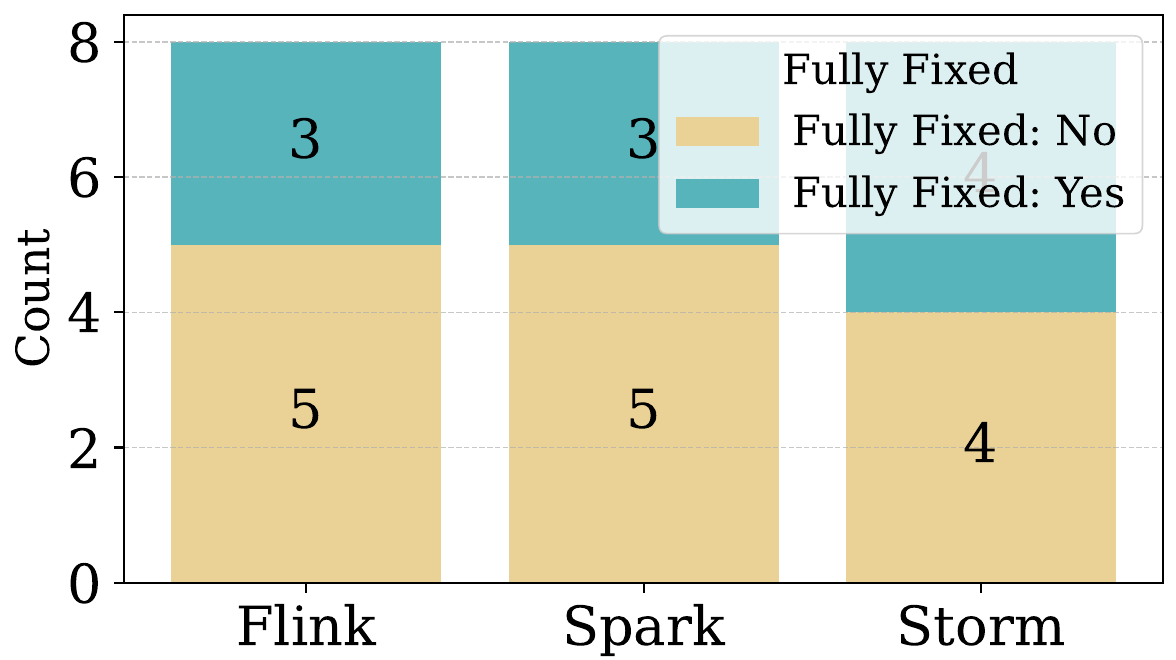}
        \vspace{-0.5cm}
        \caption{\scriptsize Distribution of fully fixed cases by SPE}
        \label{fig:fancy_fixed}
    \end{subfigure}%
    \hfill%
    % Second Subfigure
    \begin{subfigure}[t]{0.48\columnwidth}
        \centering
        \includegraphics[width=\linewidth]{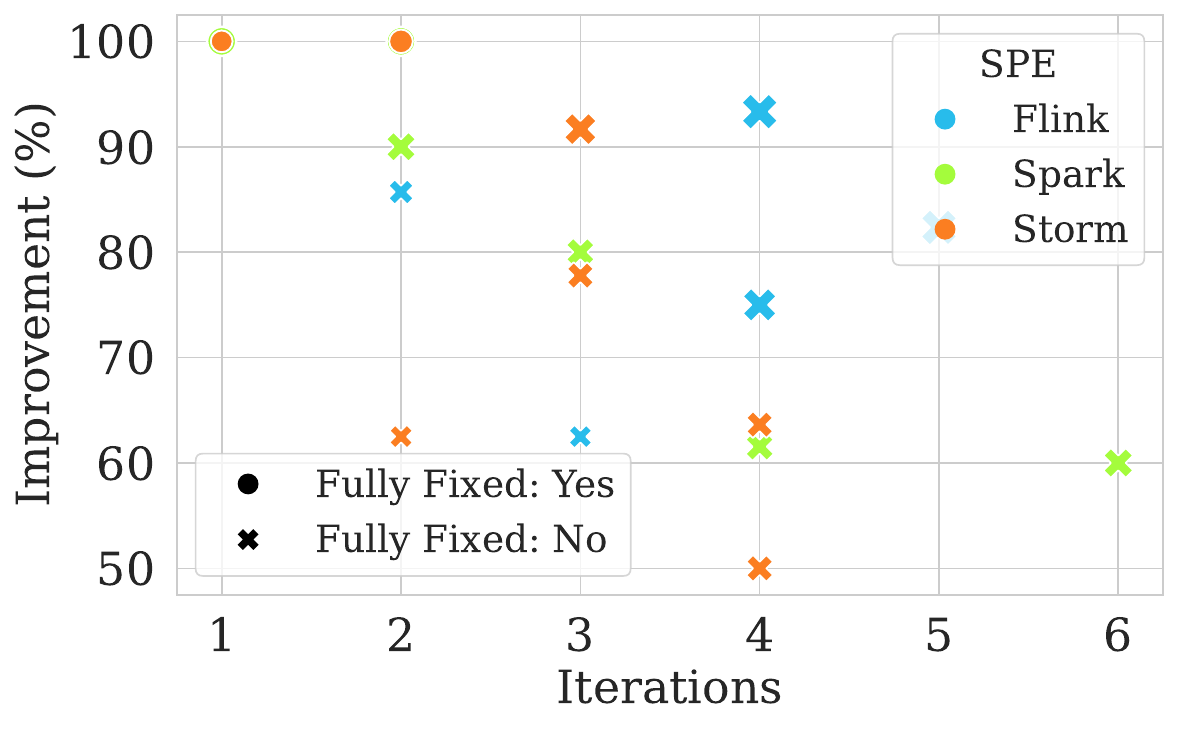}
        \vspace{-0.5cm}
        \caption{\scriptsize Iterations vs Improvement (Bubble size: Error reduction)}
        \label{fig:iteration}
    \end{subfigure}

    \vspace{0.5cm}
    \caption{Overview of fully fixed outcomes and iterative improvements across different SPEs.}
    \label{fig:fixFully}
\end{figure}

\textbf{Iteration and Improvement Relation.} Figure~\ref{fig:fixFully} presents two complementary views of improvement outcomes in all SPEs. In Figure~\ref{fig:fancy_fixed}, we compare the proportion of fully fixed solutions (upper segment of each bar) against partially fixed solutions (lower segment), which gives insight into how often each SPE achieves complete resolution of the issue. In Figure~\ref{fig:iteration}, we illustrate the number of iterations (x-axis) required to achieve a particular percentage of improvement (y-axis), with each point representing a single experimental run. The bubble size encodes the magnitude of the error reduction (i.e., the number of errors addressed), while distinct colors identify the SPE, and marker shapes indicate whether the final solution was fully fixed. Taken together, we understand from figures~\ref{fig:fancy_fixed} and \ref{fig:iteration} that elucidate which SPEs tend to yield complete fixes more consistently, that 3 or 4 iterations are typically necessary to reach high levels of improvement, and the relative difficulty of resolving errors across different SPE scenarios. 

\begin{table}
\centering
\setlength{\extrarowheight}{0pt}
% \addtolength{\extrarowheight}{\aboverulesep}
% \addtolength{\extrarowheight}{\belowrulesep}
\setlength{\aboverulesep}{0pt}
\setlength{\belowrulesep}{0pt}
\caption{Development Time and Performance Metrics by Pipeline Complexity}
\label{table:comprehensive_performance}
\resizebox{\linewidth}{!}{%
\begin{tabular}{cccccccccc} 
\toprule
\rowcolor[rgb]{0.816,0.816,0.816} {\cellcolor[rgb]{0.816,0.816,0.816}} & \multicolumn{3}{c}{~\textbf{Development Time (min)}} & \multicolumn{3}{c}{\textbf{Throughput (k events/s)}} & \multicolumn{3}{c}{\textbf{Latency P99 (ms)}} \\
\rowcolor[rgb]{0.816,0.816,0.816} \multirow{-2}{*}{{\cellcolor[rgb]{0.816,0.816,0.816}}\textbf{Complexity}} & \textbf{\textsc{ASP}} & \textbf{NiFi} & \textbf{Manual} & \textbf{\textsc{ASP}} & \textbf{NiFi} & \textbf{Manual} & \textbf{\textsc{ASP}} & \textbf{NiFi} & \textbf{Manual} \\ 
\hline
Simple & 10–20 & 10–20 & 40–90 & 150 & 138 & 153 & 12.8 & 18.3 & 10.8 \\
Medium & 15–30 & 20–40 & 70–140 & 98 & 81 & 105 & 28.1 & 40.5 & 25.4 \\
Complex & 25–45 & 45–70 & 150–300 & 72 & 60 & 78 & 55.7 & 84.7 & 54.4 \\
\bottomrule
\end{tabular}
}
\scriptsize \textit{Note}: Throughput and latency metrics were measured after any errors in the generated pipelines.
\end{table}

\textbf{Development Time: Speeding Up Production. }
The results show that \textsc{AutoStreamPipe} significantly reduces development effort. For simple pipelines, its development time (10 to 20 minutes) is similar to visual tools like NiFi. However, as task complexity increases, the advantage becomes much clearer. For complex pipelines, \textsc{AutoStreamPipe} takes only 25 to 45 minutes, making it up to 1.6$\times$ faster than NiFi (which takes 45 to 70 minutes).
This improvement is especially noticeable compared to manual coding. An expert developer might spend up to five hours (300 minutes) creating a complex pipeline, while \textsc{AutoStreamPipe} completes the same task in about 35 minutes, achieving a remarkable 6.3$\times$ reduction in development time. This speedup is due to the framework's ability to convert high-level user intent into efficient, platform-specific code, effectively addressing a key challenge mentioned in Section~\ref{sec:introduction}.

While speeding up development is a key goal, the generated pipelines must also perform well in production environments. Our analysis shows that \textsc{AutoStreamPipe} achieves a strong balance between automation and runtime efficiency.
When compared to the hand-written baseline, the pipelines produced by \textsc{AutoStreamPipe} have only a slight performance overhead. The throughput is, on average, 2 to 8 percent lower than that of the manually optimized code. Similarly, the end-to-end latency increases by 2 to 18 percent. This demonstrates that the automated generation process incurs minimal performance cost, confirming its effectiveness in challenging, real-world situations.
In contrast to the visual, low-code approach of Apache NiFi, \textsc{AutoStreamPipe} shows significant performance improvements. Across all levels of complexity, our generated pipelines achieve 15 to 20 percent higher throughput and 30 to 35 percent lower latency than those created with NiFi. This indicates that \textsc{AutoStreamPipe} not only streamlines development but also produces pipelines that are much more efficient than those from traditional low-code tools.

In summary, \textsc{AutoStreamPipe} delivers the fast development experience of a low-code platform while ensuring that pipelines maintain nearly optimal performance, comparable to that of expert manual implementation.

%%%%%%%%%%%%%%%%%%%%%%%%%%%%%%%%%%%%%%%%%%%%%%%%%%%%%%%%%%%%%%
\section{Conclusion} \label{sec:conclusion}
In this work, we introduce \textsc{AutoStreamPipe}, a novel framework for automating the generation of data stream processing pipelines employing modern LLMs. \textsc{AutoStreamPipe} directly addresses the challenges faced by domain experts who lack programming expertise but require sophisticated SP pipelines to meet their needs. 
Our approach is built on three key innovations:
\textbf{(1)} a specialized user query analyzer that performs intent detection and parameter extraction to construct tailored execution plans,
\textbf{(2)} a novel cognitive reasoning framework, which we term the \textit{Hypergraph of Thoughts}, that decomposes high-level semantic queries into executable subtasks and synthesizes pipelines across diverse stream processing engines, and
\textbf{(3)} a fault-tolerant multi-agent execution layer that coordinates specialized LLM agents to collaboratively produce and refine pipelines with error recovery mechanisms.
Our evaluation of eight diverse workloads and three SPEs demonstrates that \textsc{AutoStreamPipe} reduces error rates by 5.19$\times$ and lowers pipeline development time by 6.3$\times$ and reduces the response time on average $27\%$ compared to state-of-the-art baselines. The system consistently outperforms baselines in terms of correctness (executable rate) and stability (output consistency). We also propose the Error-Free Score (EFS) metric to assess the proportion of fully correct and executable pipelines, which offers a more comprehensive evaluation criterion than syntactic validation alone. Furthermore, the integration of persistent memory and adaptive agent orchestration enables the system to maintain contextual fidelity across complex, multi-turn interactions.
Future work will include: \textbf{(i)} support for real-time schema evolution and dynamic workloads, and \textbf{(ii)} integration with self-healing mechanisms for runtime fault adaptation.

\section*{Acknowledgement}
This article is an output of a project supported by the Recovery and Resilience Plan of the Slovak Republic under the call 'Transformation and Innovation Consortia' (project code: 09I02-03-V01-00012). The author contributes to this research as part of a consortium coordinated by InterWay, a.s.

\bibliographystyle{elsarticle-num} 
\bibliography{references}

\end{document}